\documentclass{article}
\usepackage{kr}

\usepackage{times}
\usepackage{soul}
\usepackage{url}
\usepackage[hidelinks]{hyperref}
\usepackage[utf8]{inputenc}
\usepackage[small]{caption}
\usepackage{graphicx}
\usepackage{amsmath}
\usepackage{amsthm}
\usepackage{booktabs}
\usepackage{algorithm}
\usepackage{algorithmic}
\usepackage{pdfpages}

\usepackage{epsfig}
\usepackage{amssymb}
\usepackage{multirow}
\usepackage[normalem]{ulem}
\usepackage{enumitem}
\usepackage{xcolor}
\usepackage{colortbl}
\usepackage{balance}
\useunder{\uline}{\ul}{}

\title{Case-based Reasoning Augmented Large Language Model Framework for Decision Making in Realistic Safety-Critical Driving Scenarios}

\author{%
Wenbin Gan\and
Minh-Son Dao\and
Koji Zettsu\\
\affiliations
Big Data Integration Research Center,\\
National Institute of Information and Communications Technology (NICT),\\ Tokyo, Japan\\
\emails
wenbingan@nict.go.jp
}

\begin{document}

\maketitle

\begin{abstract}
  Driving in safety-critical scenarios requires quick, context-aware decision-making grounded in both situational understanding and experiential reasoning. Large Language Models (LLMs), with their powerful general-purpose reasoning capabilities, offer a promising foundation for such decision-making. However, their direct application to autonomous driving remains limited due to challenges in domain adaptation, contextual grounding, and the lack of experiential knowledge needed to make reliable and interpretable decisions in dynamic, high-risk environments. To address this gap, this paper presents a Case-Based Reasoning Augmented Large Language Model (CBR-LLM) framework for evasive maneuver decision-making in complex risk scenarios. Our approach integrates semantic scene understanding from dashcam video inputs with the retrieval of relevant past driving cases, enabling LLMs to generate maneuver recommendations that are both context-sensitive and human-aligned. Experiments across multiple open-source LLMs show that our framework improves decision accuracy, justification quality, and alignment with human expert behavior. Risk-aware prompting strategies further enhance performance across diverse risk types, while similarity-based case retrieval consistently outperforms random sampling in guiding in-context learning. Case studies further demonstrate the  framework's robustness in challenging real-world conditions, underscoring its potential as an adaptive and trustworthy decision-support tool for intelligent driving systems.   
\end{abstract}

\section{Introduction}

Ensuring safety in complex and multifaceted driving 
scenarios remains a paramount concern in the advancement of intelligent driving systems \cite{chib2023recent,li2024large,garikapati2024autonomous}.
Autonomous vehicles (AVs) and advanced driving assistance systems (ADASs) must not only respond to immediate risks, but also consider dynamic factors to enable contextually appropriate and human-like decision-making. 
Given the significant human toll of annual traffic crashes \cite{world2019global} and the the increasing presence of AVs in complex real-world scenarios \cite{garikapati2024autonomous}, the necessity of building robust and trustworthy intelligent systems has become increasingly apparent.

\begin{figure}[t]
\centerline{\includegraphics[width=0.9\linewidth]{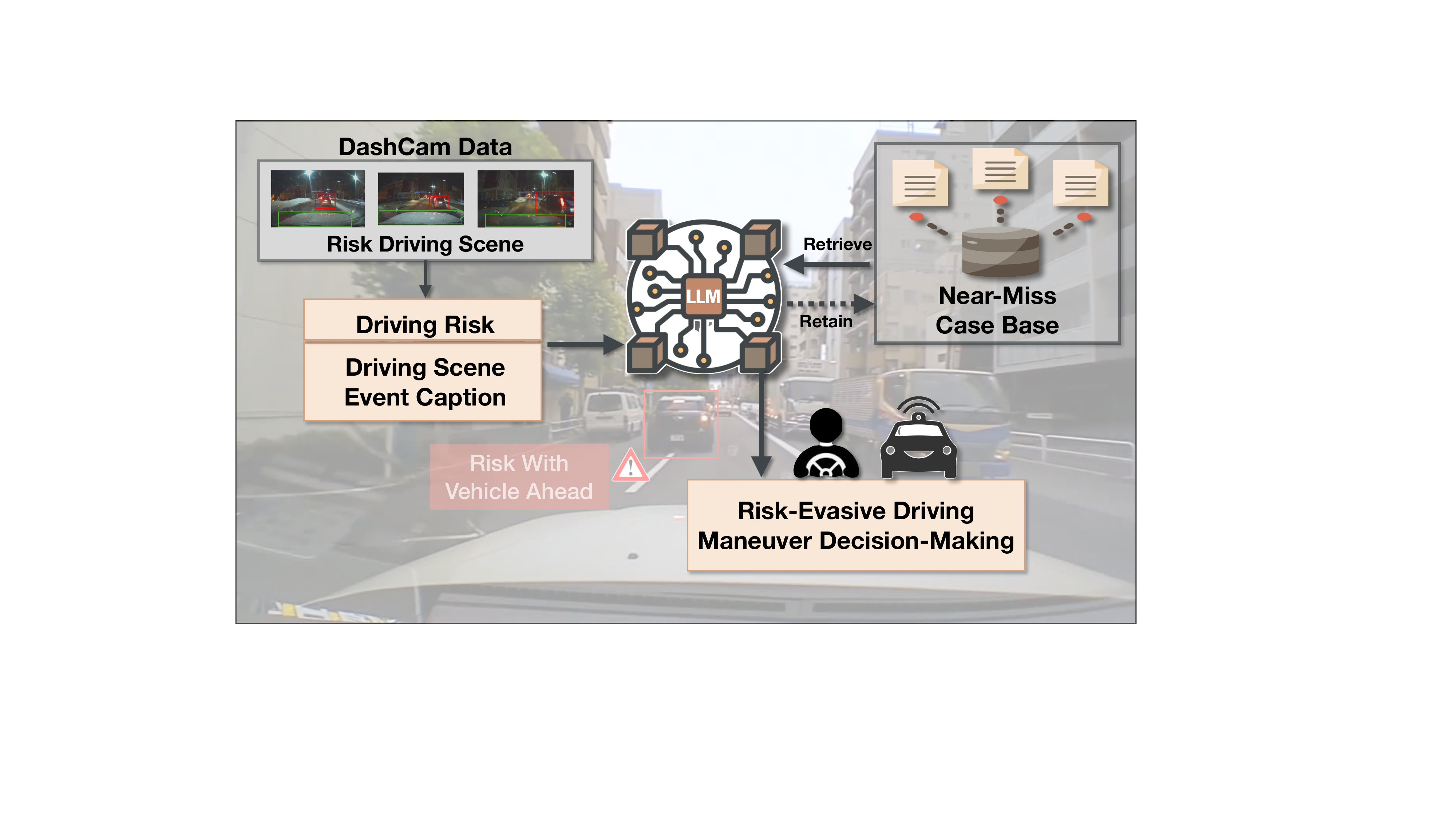}}
\caption{We present CBR-LLM, a case-based reasoning augmented LLM framework for driving decision-making in risk scenarios. It takes as input the DashCam video data and outputs the recommended risk-evasive maneuver for specific risk scenario (e.g., ``emergency braking and evasive steering right'' to avoid the risk of collision with the vehicle ahead).}
\label{introduction_graph}
\vspace{-15 pt}
\end{figure}

While significant advancements have been achieved in perception, planning, and control modules \cite{chib2023recent}, the challenge of risk-sensitive decision-making in various safety-critical driving scenarios (SCDSs) remains a fundamental barrier to the deployment of trustworthy systems \cite{shi2024scvlm,zhou2024safedrive,zhao2025saca,gan2024drive,gan2022open}.
SCDSs are high-risk on-road situations where driver error or inadequate response (human or autonomous) can lead to accidents or injuries, often involving unexpected events, demanding conditions, or complex interactions \cite{ding2023survey,gan2024drive}.

While conventional rule-based systems \cite{li2021risk} (trigger action using predefined rules) and end-to-end data-driven systems \cite{chen2024end} (feed sensor data directly to vehicle control variables) have demonstrated promising performance to provide maneuver assistance in structured scenarios, their generalization capacity is often constrained by distributional biases \cite{zhao2025saca}. Specifically, such models are typically trained on over-represented normal-driving data, leading to diminished performance when faced with safety-critical, long-tail cases \cite{shi2024scvlm,shao2024lmdrive}. Moreover, the black-box nature of these data-driven models hinders interpretability and fails to provide justifiable reasoning behind maneuvers \cite{zhou2024safedrive}, both of which are critical for safety validation and human trust.
In contrast to these systems, expert human drivers typically leverage their experiential knowledge to execute analogous maneuvers when reacting to SCDS \cite{gan2022open,gan2023procedural}.
These bring us to a key question: \textbf{\textit{How can we help autonomous systems make risk-sensitive decisions in the SCDSs that are not only contextually appropriate but also explainable and grounded in experience?}}
From the perspective of Knowledge Representation and Reasoning (KR), this gap highlights the need for method that explicitly encodes situational knowledge and supports structured, semantically grounded inference for safety-critical decision-making.

\begin{figure}[t]
\centerline{\includegraphics[width=0.9\linewidth]{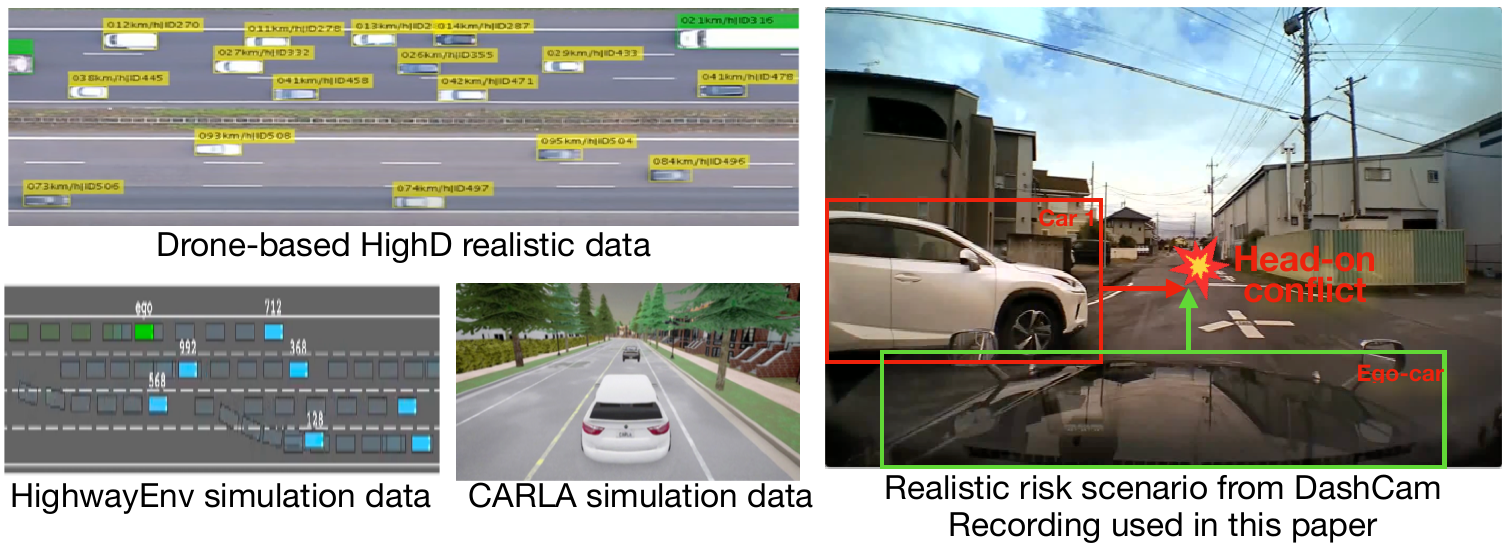}}
\caption{Different types of data used for driving decision-making studies.}
\label{realistic_SCDS}
\vspace{-15 pt}
\end{figure}

Recent progress in LLMs has unveiled novel potential for emulating human-like high-level reasoning and commonsense-grounded decision-making in the field of autonomous driving \cite{li2024large,xu2024drivegpt4}. 
Leveraging the LLMs-driven reasoning, existing studies have demonstrated enhanced cognitive capabilities of driving agents \cite{jin2023adapt,xu2024drivegpt4,shao2024lmdrive,fu2024drive,shi2024scvlm,cai2024driving}.
However, their deployment in SCDSs remains limited due to concerns around hallucination, overconfidence, and a lack of embedded domain-specific knowledge.
Despite a few pioneer studies explored LLMs for driving assistance in SCDSs, most researches apply LLMs on simple simulation driving environments \cite{shao2024lmdrive,wen2023dilu} or fine-grained drone-based Bird's Eye View (BEV) context \cite{zhou2024safedrive} with full observability of all agents and their trajectories, which is rarely the case in reality. While systems utilizing an egocentric viewpoint for input (e.g., Dashcam videos shown in Figure \ref{realistic_SCDS}) offer greater scalability and ease of deployment, especially when the goal is to build realistic and human-like decision-making systems in real-world driving environments. However, the increased complexity and dynamic nature of real-world driving scenarios with partial observability pose additional challenge for LLM-based agents in ensuring safe decision-making.

To address these challenges, this paper proposes a Case-Based Reasoning Augmented LLM (CBR-LLM) framework for human-like driving decision-making in realistic SCDSs, leveraging DashCam video data. CBR-LLM strategically integrates the robust generalization and reasoning capabilities of LLMs with the structured recall and adaptive mechanisms for experiential knowledge offered by CBR. 
Our approach is inspired by cognitive science on human decision-making, where human drivers often rely on prior experiences to navigate new, yet analogous, risk scenarios. 
By emulating this cognitive paradigm, the proposed CBR-LLM framework operates by retrieving relevant historical SCDS cases from a curated memory bank and adapts them to current situational contexts using LLM-based reasoning. By embedding CBR within the LLM-driven reasoning loop, we aim to enhance the model's access to contextually rich, risk-related knowledge without compromising the flexibility and generalization strengths that characterize LLMs.

The proposed CBR-LLM framework is designed to operate in realistic SCDSs, as shown in Figure \ref{introduction_graph}. It first transforms the risk scenarios into the scene event captions and risk types, and then retrieves similar historical SCDS cases based on current scene information, adapts those cases through LLM reasoning, and outputs evasive-maneuver recommendations that are both context-aware and risk-sensitive. This design bridges the gap between data-driven generalization and experience-based specificity, allowing the system to make decisions that align more closely with human-like driving behavior.
To validate the framework, we conduct experiments using real-world driving datasets containing diverse SCDSs. 
Our analysis focuses not only on decision alignment (i.e., correspondence with expert driving decision) but also on the reasoning and interpretability of various open-source LLMs in safety-critical driving contexts.
The results show that the CBR-LLM framework effectively integrates case-based retrieval with LLM reasoning, enabling the generation of contextually appropriate maneuver decisions. Notably, the system provides coherent, human-understandable justifications, enhancing transparency and trust.
Among the tested models, LLaMA3.3-70B achieved the highest performance (0.9412), indicating strong alignment with human-like decision-making in risk situations. These findings highlight the potential of the CBR-LLM framework for robust, adaptive, and interpretable decision-making in real-world driving applications. Its ability to continuously learn from evolving naturalistic driving data further underscores its practical applicability for deployment in safety-critical driving systems.

The main contributions of this work are as follows:
\begin{itemize}[leftmargin=15pt,itemsep=2pt,topsep=0pt,parsep=0pt]
\item We propose a CBR-LLM framework that augments LLM-based decision-making with structured, experiential knowledge via Case-Based Reasoning, bridging classical symbolic memory with modern neural reasoning, advancing the KR goal of combining declarative representations with flexible, data-driven inference.
\item We design an integrated decision-making pipeline leveraging historical risk driving scenarios for context-aware maneuver recommendation in safety-critical situations.
\item We evaluate our framework alongside popular LLMs on real-world datasets, demonstrating effective maneuver recommendations aligned with expert driving decisions.
\end{itemize}

\section{Related Work}

Reasoning and decision-making are critical for intelligent driving systems since they assist a vehicle (driver) in navigating complex and dynamic scenarios \cite{li2021risk,cai2024driving,shao2024lmdrive,zhou2024safedrive}. 
Existing approaches to driving reasoning and decision-making often rely on rule-based systems grounded in statistical learning or end-to-end neural models. 
Li et al. proposed a risk assessment based decision-making method using  rule-based control logics, where take over time and braking control strategy were proposed to avoid collision \cite{li2021risk}. 
Compared to this, end-to-end driving decision making presents more flexibility to various routine scenarios. Chen et al. summarized existing studies on the end-to-end decision making in autonomous driving. Imitation learning and reinforcement learning are the two widely used paradigms to map the sensor data directly into driving control signals \cite{chen2024end}. While these methods have demonstrated good performance in routine scenarios, they typically lack interpretability and robustness in rare or high-risk SCDSs \cite{zhou2024safedrive,zhao2025saca}. 
To cope with the SCDSs, 
Shi et al. proposed a method for better understanding SCDSs from driving videos \cite{shi2024scvlm}, providing valuable insights for decision-making in such risk scenarios. Gan et al. proposed an open CBR based evolving framework for driving assistance in risk scenarios \cite{gan2022open}. 

Recent interest in using LLMs for high-level reasoning tasks has led to their application into the decision-making for autonomous driving \cite{li2024large}. 
Xu et al. used multimodal LLMs to respond to human inquiries about the vehicle actions, providing the autonomous driving interpretation \cite{xu2024drivegpt4}. Shao et al. introduced the LMDrive, a language-guided autonomous driving framework with decision-making and execution in a closed-loop within a driving simulator \cite{shao2024lmdrive}. Wang et al.  \cite{wang2023drivemlm} joined user instructions into the decision-making process of LLMs. These studies highlight the great potential of LLMs for autonomous driving.
To enhance the reliability of reasoning and decision-making with LLMs, some studies explored to use external memory to store common-sense knowledge, ensuring the knowledge driven decision-making.
Fu et al. proposed the DiLu framework to leverage LLMs to perform decision-making based on external knowledge, and let vehicles drive like a human \cite{fu2024drive,wen2023dilu}. 
Cai et al. integrated LLMs with retrieval-augmented reasoning for driving decision-making, ensuring traffic rule and guideline-adherent decision outputs and seamless adaptation to different regions \cite{cai2024driving}. 
Zhou et al. conducted knowledge- and data-driven decision-making in risk scenarios with LLMs \cite{zhou2024safedrive}, showing its effectiveness in enhancing safety.
These findings offer valuable insights for the current study for knowledge- and data-driven decision-making with LLMs.

Despite their great potential, most existing studies apply LLMs to simplified simulated driving environments \cite{shao2024lmdrive,wen2023dilu} or rely on drone-based BEV representations \cite{zhou2024safedrive}. To move toward realistic and human-aligned decision-making, it is essential to develop systems that operate in real-world driving environments from an egocentric perspective.

\begin{figure}[t]
\centerline{\includegraphics[width=0.8\linewidth]{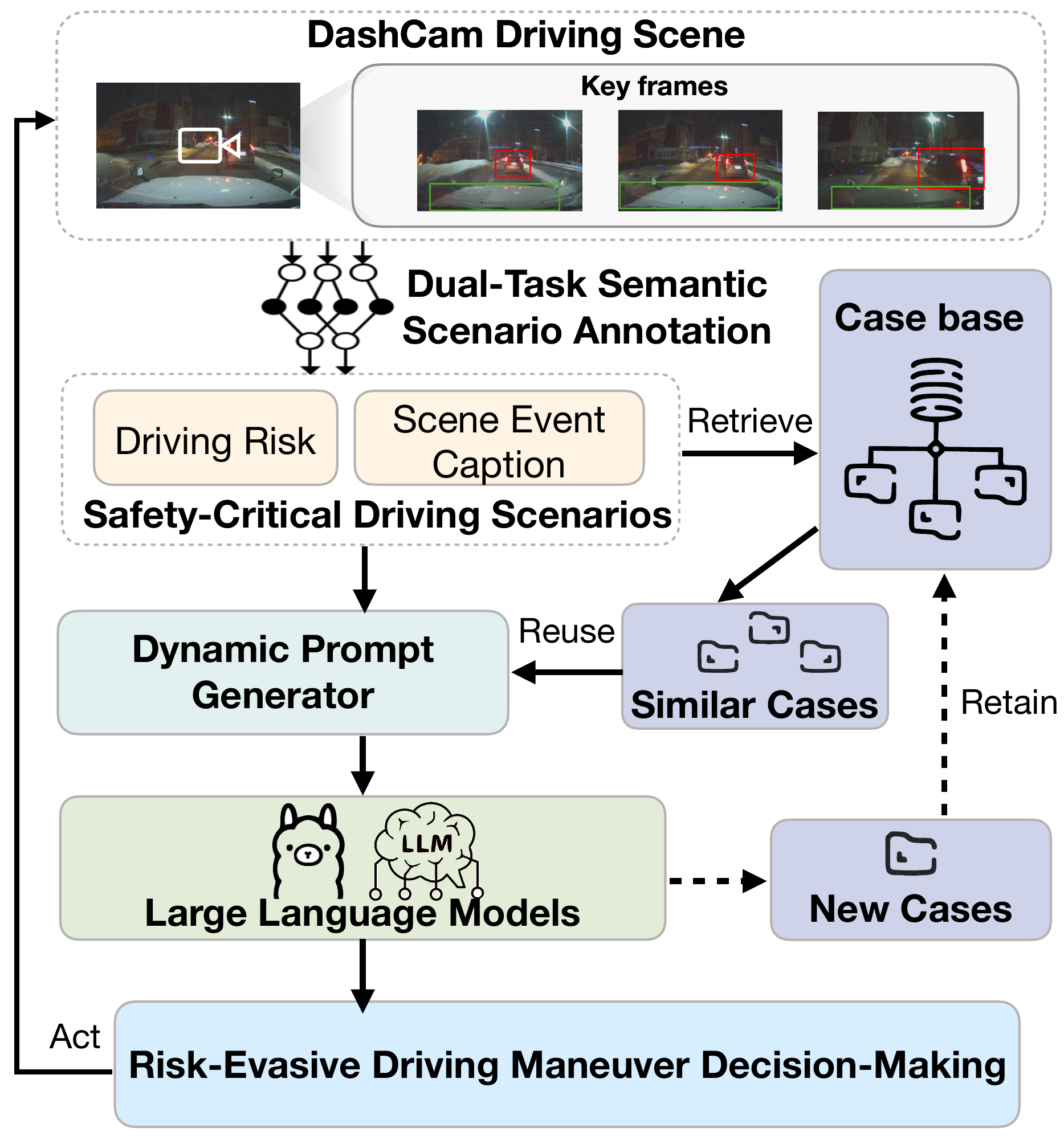}}
\caption{Overview of our CBR-LLM framework for driving maneuver decision-making in safety-critical scenarios.}
\label{framework}
\vspace{-15 pt}
\end{figure}

\section{Methodology}
Figure \ref{framework} shows the overall CBR-LLM framework for the driving maneuver decision-making in the SCDSs. 
We build an evolving case base to store the experiences and knowledge of human drivers to successfully cope with diverse risk driving scenarios. These experiences are stored as cases with rich information to indicate the event context, risk categories, risk-evasive maneuver taken and some justification. 
Our framework takes the Dashcam driving video data as input, combines the strengths of CBR and LLMs for augmented reasoning, and outputs the adaptive, human-like driving maneuvers in safety-critical scenarios. 

It solves the task in three stages: initially, a dual-task scenario understanding model is applied to extract risk event-related information from the Dashcam videos. It not only predicts the detailed driving risk (e.g., conflict with the vehicle ahead) in the current situation, but also generates a comprehensive risk event caption that captures the spatial-temporal dynamics of risk formation. Hence it is an event-based paradigm rather than a frame-based one. Subsequently, CBR is conducted using this risk event information to retrieve from the case base to obtain similar risk situations. Finally, 
the LLM reasoning module integrates the current risk event information and the recalled similar cases into the dynamic prompt generator to generate a comprehensive prompt for the adopted LLM. Chain-of-Thought reasoning is then conducted by the LLM to generate the contextually appropriate, risk-evasive decision-making. 
These decisions are then provided to an AV driving agent for vehicle control to mitigate the identified risk, or to assist human drivers in a driver-in-the-loop conditionally automated system.

\subsection{Semantic Annotation of SCDSs}
\label{Semantic_Annotation_SCDSs}
To enable a deeper understanding of SCDSs, we adopt a dual-task semantic analysis model that extracts risk-related event information directly from dashcam videos. This model performs two complementary tasks: predicting detailed driving risks and generating natural language descriptions that capture the spatial-temporal context of those risks. 

Our method is built upon the model of \cite{shi2024scvlm}, we adapt their pipeline by modifying the risk prediction module using a two-stage prediction model in previous work \cite{dao2023mm,gan2025smart} trained with different bitch of data in the same dataset in our experiment
(Refer to original papers for further details).
The first task focuses on risk assessment using a two-stage prediction pipeline. In the initial stage, the model determines whether the current driving scene is safe or presents a potential risk. If a risk is detected, the second stage classifies the event into one of seven predefined risk types (such as conflict with vehicle ahead). This layered structure ensures that both the presence and the nature of the risk are accurately captured.
The second task is designed to generate comprehensive risk event captions. A dedicated output head in the model integrates multiple semantic components, including risk category, conflict type, and driving context. These features are encoded and passed to a LLM (we use DeepSeek-R1 \cite{guo2025deepseek}), which synthesizes a narrative-style description of the event. The resulting caption captures not only the immediate risk but also the spatial-temporal dynamics (see the examples in Figure \ref{example1} and \ref{example2}).

By combining structured risk classification with free-form language generation, the semantic annotations serve as proper inputs for leveraging LLMs for downstream tasks of risk-sensitive decision-making. 
To more accurately evaluate our further CBR-LLM framework, we manually check all the semantic annotations of dashcam videos to ensure the correctness. Moreover, we also label the SCDSs with the actual (ground-truth) evasive maneuvers executed by human drivers to successfully mitigate the identified risks (risk types and maneuver set are shown in Table 1). Note that the evasive maneuver attach with one video is one in the maneuver set of Table 1.
Note that each video is associated with a single evasive maneuver from the maneuver set listed in Table 1.

\begin{table}[t]
\centering
\label{Risk_categories}
\caption{Risk types of SCDSs and evasive driving maneuvers.}
\vspace{-10 pt}
\resizebox{\linewidth}{!}{ %
\begin{tabular}{@{}l|l@{}}
\toprule
\textbf{Safety-Critical Driving Scenarios}      & \textbf{Evasive Driving Maneuvers}                     \\ \midrule
Conflict with Vehicle Ahead            & Emergency Braking                            \\
Conflict with Pedestrian               & Evasive Steering Left                        \\
Conflict with Adjacent Vehicle         & Evasive Steering Right                       \\
Conflict with Merging Vehicle          & Sudden Acceleration                          \\
Conflict with Oncoming Vehicle         & Emergency Braking and Evasive Steering Left  \\
Head-on Conflict                       & Emergency Braking and Evasive Steering Right \\
Conflict with Opposite Turning Vehicle & Acceleration and Evasive Steering Left       \\
                                       & Acceleration and Evasive Steering Right      \\ \bottomrule
\end{tabular}
}
\vspace{-20 pt}
\end{table}

\subsection{CBR Augmented Human-Experience Recall}
To support context-aware and experience-driven decision-making, our framework incorporates a CBR module that selectively recalls semantically similar SCDSs based on the current risk scene. 
Drawing on human decision-making's reliance on prior experiences, the CBR module enables the system to recall relevant historical knowledge from a case base and apply it to real-time driving contexts.

We initially build a case base by selecting a few resolved SCDSs with manually pre-defined risk analysis items and the executed evasive maneuvers. Each cases is with the following form: 

\textit{$\langle$ \textbf{Key}: Risk type, Scene event caption; \textbf{Value}: risk analysis results [road context, other car position, other car action, Event context, ego-car evasive maneuver, ego-car maneuver justification]$\rangle$}. 

The key part is used in similarity calculation for retrieving cases while the value part stores the decision-making logic for risk resolvation. 

When a new risk scenario is encountered, the system analyzes key scene-level features, such as risk types and scene event caption, to identify similar cases stored in the memory.
We use an open embedding model ``nomic-embed-text'' to embed the event captions into vectors and attach them to the corresponding cases. The similarity calculation of vectors is used to find the most resemble SCDSs for the current situation. During the similar case retrieving, we constraint the recalling process within the same risk types, as risk events in the same categories often share similar risk driving patterns and are more effective to adapt to current situation. 
The retrieved cases are integrated into the decision-making process, enriching the framework's reasoning with precedent-based knowledge. This allows the system to better generalize in unfamiliar or uncertain situations and improves decision robustness by grounding responses in real-world driving decisions.

Moreover, the case base is designed to evolve. As the system encounters new scenarios and observes their outcomes, it incrementally updates the case base with newly analyzed experiences. This continual refinement ensures that the case base remains diverse, up-to-date, and reflective of complex, real-world SCDSs. As a result, the framework supports a dynamic form of learning that aligns closely with how human drivers improve through experience over time.

\subsection{LLMs-based Risk-Sensitive Decision-Making}

\begin{figure}[t]
\centerline{\includegraphics[width=0.8\linewidth]{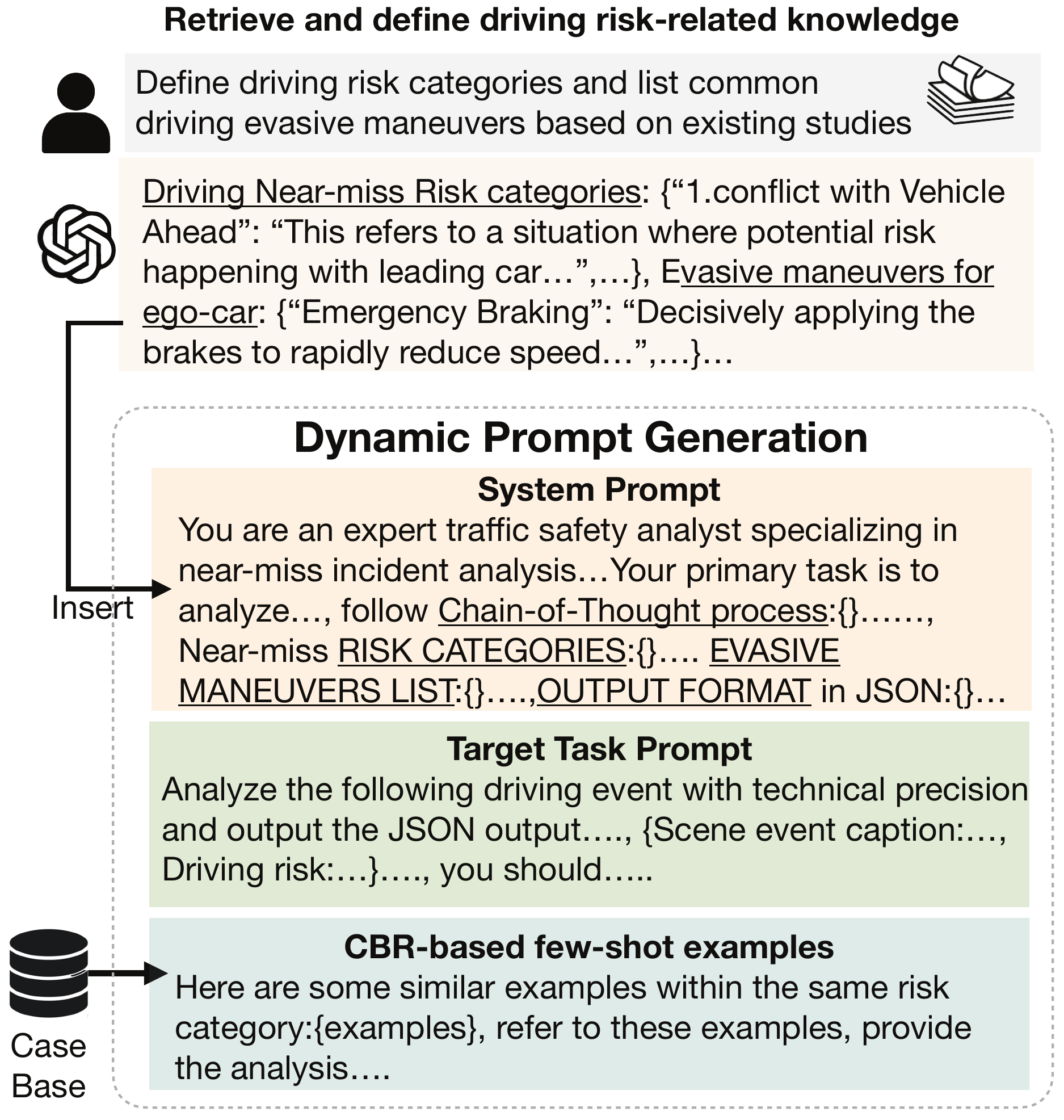}}
\caption{The process of dynamic prompt generation.}
\label{dynamic_prompt}
\vspace{-15 pt}
\end{figure}

We use a dynamic prompt generator for creating the input prompts to feed into the LLMs. As shown in Figure \ref{dynamic_prompt}, we first retrieve and define the driving risk related knowledge using the OpenAI's ChatGPT based on existing studies. It provides foundational information, such as the common near-miss risk categories and their definition, and the common evasive maneuvers to resolve risks. This information is then polished and inserted into the prompt generator for creating the system prompt. Here system prompt is shared by all the following reasoning process, it defines the LLMs' role and technical analysis requirements, emphasizes the primary task for risk evasive decision-making, provides the accurate lists and definitions of SCDS risk categories and evasive maneuver list, as well as gives the expected output format. The LLMs are also asked to follow a carefully designed Chain-of-Thought (CoT) process introduced in \cite{wei2022chain}: it first conducts an overview of the general road context, and analyzes the risk-evolved other car's position and action in a spatial-temporal inference manner, and summarizes a clear event context, and finally infers the evasive maneuver based on all the obtained information, and gives its justification why this maneuver is recommended. With CoT, the LLMs conduct its reasoning step by step and then gives the final decision.  

Apart from the system prompt, it also have two dynamic parts for generating the target task prompt and the CBR-based few-shot examples. When a SCDS is detected, the driving risk and the scene event caption are filled into the task prompt, and 
CBR is conducted based on the current event information to recall similar cases in case base.
Similar cases in the same risk categories are integrated as few-shot examples to assist in the reasoning process. 
After the dynamic prompt generation, it is finally feeds into the LLMs for decision-making in the current SCDS. 
The LLM's output is structured in JSON format, containing rich analysis information, as illustrated in the examples of Figure \ref{example1} and \ref{example2}.
This structured output ensures the recommended maneuvers are both explainable and safety-sensitive.

\section{Experiments}

\subsection{Dataset and SCDS Event Processing}
Real-world naturalistic near-miss driving dataset\footnote{\url{https://web.tuat.ac.jp/~smrc/en/about.html}} was used in this study.
This dataset is a collection of dashcam recordings sourced from the Tokyo University of Agriculture and Technology. It comprises over 220,000 video segments, each documenting a near-accident safety-critical event and accompanied by descriptive tags. Each video spans roughly 15 seconds, capturing the 10 seconds leading up to the critical ``trigger'' point and the subsequent 5 seconds. Due to copyright constraints, we select  a subset of 1000 samples. To evaluate our framework on diverse SCDS events, we balance the numbers of videos in these seven risk types (around 143 for each). 

\begin{table}[t]
\centering
\label{Dataset_statistics}
\caption{Dataset statistics.}
\vspace{-10 pt}
\resizebox{\linewidth}{!}{ %
\begin{tabular}{l|c|lll}
\hline
\multicolumn{1}{c|}{\multirow{2}{*}{Risk Types}} & \multirow{2}{*}{\# of Events} & \multicolumn{3}{c}{Event Caption Length}                                      \\ \cline{3-5} 
\multicolumn{1}{c|}{}                               &                               & \multicolumn{1}{c}{min} & \multicolumn{1}{c}{mean} & \multicolumn{1}{c}{max} \\ \hline
Conflict with Vehicle Ahead                         & 143                           & 273                     & 455.58                   & 844                     \\
Conflict with Pedestrian                            & 142                           & 286                     & 423.42                   & 729                     \\
Conflict with Adjacent Vehicle                      & 143                           & 287                     & 438.99                   & 895                     \\
Conflict with Merging Vehicle                       & 143                           & 115                     & 218.22                   & 442                     \\
Conflict with Oncoming Vehicle                      & 143                           & 203                     & 384.35                   & 752                     \\
Head-on Conflict                                    & 143                           & 264                     & 426.51                   & 846                     \\
Conflict with Opposite Turning Vehicle              & 143                           & 265                     & 398.31                   & 766                     \\ \hline
\end{tabular}
}
\vspace{-15 pt}
\end{table}

Using the semantic annotation method in Section \ref{Semantic_Annotation_SCDSs}, we obtain the  
risk analysis and event caption information for these videos, as shown in Table 2.
Examples of data are shown in Figure \ref{example1} and \ref{example2}. We carefully review all the information for these data to assure the correctness of semantic annotation, thereby ensuring a reliable evaluation of the proposed CBR-LLM framework.

\begin{table*}[ht]
\setlength\tabcolsep{8pt}
\centering
\label{risk_aware_unaware}
\caption{Performance comparisons on risk-aware and risk-unaware evasive maneuver recommendation.}
\vspace{-10 pt}
\resizebox{1.\linewidth}{!}{ %
\begin{tabular}{@{}lccccccccc@{}}
\toprule
\multicolumn{1}{c|}{\multirow{2}{*}{\textbf{Model}}} & \multicolumn{4}{c|}{\textbf{Event Context}}                                                                                                                                                & \multicolumn{1}{c|}{\textbf{Ego-car Evasive Maneuver}}  & \multicolumn{4}{c}{\textbf{Ego-car Maneuver Justification}}                                                                                                            \\ \cmidrule(l){2-10} 
\multicolumn{1}{c|}{}                       & \multicolumn{1}{c|}{BLEU4}                 & \multicolumn{1}{c|}{METEOR}                & \multicolumn{1}{c|}{ROUGE\_L}              & \multicolumn{1}{c|}{CIDEr}                 & \multicolumn{1}{c|}{Micro-Accuracy}  & \multicolumn{1}{c|}{BLEU4}                & \multicolumn{1}{c|}{METEOR}                & \multicolumn{1}{c|}{ROUGE\_L}              & CIDEr                 \\ \midrule
\multicolumn{10}{c}{\textbf{Risk-Unaware Results}}                                                                                                                                                                                                                                                                                                                                                                                             \\ \midrule
\multicolumn{1}{l|}{Mistral 7B}            & \multicolumn{1}{c|}{14.64 (1.07)}          & \multicolumn{1}{c|}{43.13 (1.83)}          & \multicolumn{1}{c|}{41.45 (1.18)}          & \multicolumn{1}{c|}{27.88 (0.90)}          & \multicolumn{1}{c|}{0.845}          & \multicolumn{1}{c|}{5.01 (0.38)}          & \multicolumn{1}{c|}{24.45 (1.29)}          & \multicolumn{1}{c|}{{\ul 37.41 2.30}}      & 17.80 (0.74)          \\ \midrule
\multicolumn{1}{l|}{Mistral-Nemo 12B}      & \multicolumn{1}{c|}{8.89 (0.81)}           & \multicolumn{1}{c|}{32.70 (1.59)}          & \multicolumn{1}{c|}{42.60 (1.42)}          & \multicolumn{1}{c|}{22.51 (0.98)}          & \multicolumn{1}{c|}{0.850}          & \multicolumn{1}{c|}{4.60 (0.34)}          & \multicolumn{1}{c|}{23.31 (1.34)}          & \multicolumn{1}{c|}{33.74 (1.88)}          & 14.92 (0.65)          \\ \midrule
\multicolumn{1}{l|}{Mixtral 8x7B}          & \multicolumn{1}{c|}{14.95 (1.14)}          & \multicolumn{1}{c|}{42.50 (1.84)}          & \multicolumn{1}{c|}{42.26 (1.19)}          & \multicolumn{1}{c|}{28.24 (0.98)}          & \multicolumn{1}{c|}{0.867}          & \multicolumn{1}{c|}{5.92 (0.35)}          & \multicolumn{1}{c|}{27.68 (1.31)}          & \multicolumn{1}{c|}{32.90 (1.38)}          & 17.35 (0.53)          \\ \midrule
\rowcolor{gray!10} \multicolumn{1}{l|}{Phi4 14B}              & \multicolumn{1}{c|}{14.11 (0.99)}          & \multicolumn{1}{c|}{44.61 (1.60)}          & \multicolumn{1}{c|}{38.82 (1.10)}          & \multicolumn{1}{c|}{27.72 (0.79)}          & \multicolumn{1}{c|}{0.894}          & \multicolumn{1}{c|}{{\ul 9.16 0.69}}      & \multicolumn{1}{c|}{\textbf{38.20 (1.62)}} & \multicolumn{1}{c|}{34.86 (1.46)}          & {\ul 21.18 0.70}      \\ \midrule
\rowcolor{gray!10} \multicolumn{1}{l|}{Deepseek-R1 14B}       & \multicolumn{1}{c|}{\textbf{17.45 (1.59)}} & \multicolumn{1}{c|}{{\ul 46.06 2.08}}      & \multicolumn{1}{c|}{{\ul 44.32 1.44}}      & \multicolumn{1}{c|}{\textbf{30.55 (1.25)}} & \multicolumn{1}{c|}{0.913}          & \multicolumn{1}{c|}{7.96 (0.80)}          & \multicolumn{1}{c|}{31.68 (1.87)}          & \multicolumn{1}{c|}{\textbf{38.52 (1.80)}} & 20.01 (0.93)          \\ \midrule
\rowcolor{gray!10} \multicolumn{1}{l|}{Gemma2 27B}            & \multicolumn{1}{c|}{15.57(1.2)}            & \multicolumn{1}{c|}{42.98 (1.70)}          & \multicolumn{1}{c|}{43.90 (1.37)}          & \multicolumn{1}{c|}{29.78 (1.05)}          & \multicolumn{1}{c|}{0.893}          & \multicolumn{1}{c|}{7.85 (0.55)}          & \multicolumn{1}{c|}{{\ul 36.21 1.40}}      & \multicolumn{1}{c|}{31.81 (1.30)}          & 20.12 (0.61)          \\ \midrule
\rowcolor{gray!10} \multicolumn{1}{l|}{Qwen2.5 32B}           & \multicolumn{1}{c|}{{\ul 16.811.42}}       & \multicolumn{1}{c|}{\textbf{47.10 (1.78)}} & \multicolumn{1}{c|}{41.77 (1.38)}          & \multicolumn{1}{c|}{30.11 (1.12)}          & \multicolumn{1}{c|}{0.875}          & \multicolumn{1}{c|}{8.85 (0.64)}          & \multicolumn{1}{c|}{1.78 (1.38)}           & \multicolumn{1}{c|}{1.38 (1.25)}           & \textbf{30.11 (0.65)} \\ \midrule
\rowcolor{gray!10} \multicolumn{1}{l|}{Llama3.3 70B}          & \multicolumn{1}{c|}{16.08 (1.54)}          & \multicolumn{1}{c|}{41.33 (1.96)}          & \multicolumn{1}{c|}{\textbf{46.24 (1.67)}} & \multicolumn{1}{c|}{{\ul 30.27 1.40}}      & \multicolumn{1}{c|}{\textbf{0.937}} & \multicolumn{1}{c|}{\textbf{9.17 (0.96)}} & \multicolumn{1}{c|}{31.92 (1.90)}          & \multicolumn{1}{c|}{32.87 (1.81)}          & 20.11 (1.05)          \\ \midrule
\multicolumn{10}{c}{\textbf{Risk-Aware Results}}                                                                                                                                                                                                                                                                                                                                                                                             \\ \midrule
\rowcolor{gray!10} \multicolumn{1}{l|}{Phi4 14B}              & \multicolumn{1}{c|}{12.46(0.70)}           & \multicolumn{1}{c|}{48.98(1.11)}           & \multicolumn{1}{c|}{33.79(0.94)}           & \multicolumn{1}{c|}{28.19(0.64)}           & \multicolumn{1}{c|}{0.885($\downarrow$) }          & \multicolumn{1}{c|}{5.54(0.28)}           & \multicolumn{1}{c|}{39.17(1.07)}           & \multicolumn{1}{c|}{22.22(0.56)}           & 18.87(0.37)           \\ \midrule
\rowcolor{gray!10} \multicolumn{1}{l|}{Deepseek-R1 14B}       & \multicolumn{1}{c|}{{\ul 18.44(1.40)}}     & \multicolumn{1}{c|}{\textbf{51.58(1.81)}}  & \multicolumn{1}{c|}{\textbf{42.26(1.43)}}  & \multicolumn{1}{c|}{{\ul 32.23(1.16)}}     & \multicolumn{1}{c|}{0.914($\uparrow$)}          & \multicolumn{1}{c|}{{\ul 8.96(0.71)}}     & \multicolumn{1}{c|}{38.37(1.77)}           & \multicolumn{1}{c|}{\textbf{32.89(1.36)}}  & \textbf{21.06(0.75)}  \\ \midrule
\rowcolor{gray!10} \multicolumn{1}{l|}{Gemma2 27B}            & \multicolumn{1}{c|}{16.08(1.13)}           & \multicolumn{1}{c|}{45.67(1.53)}           & \multicolumn{1}{c|}{41.46(1.17)}           & \multicolumn{1}{c|}{29.87(0.96)}           & \multicolumn{1}{c|}{{\ul 0.918($\uparrow$)}}    & \multicolumn{1}{c|}{5.85(0.33)}           & \multicolumn{1}{c|}{35.32(1.01)}           & \multicolumn{1}{c|}{24.66(0.72)}           & 18.45(0.45)           \\ \midrule
\rowcolor{gray!10} \multicolumn{1}{l|}{Qwen2.5 32B}           & \multicolumn{1}{c|}{16.84(1.20)}           & \multicolumn{1}{c|}{50.88(1.42)}           & \multicolumn{1}{c|}{40.12(1.24)}           & \multicolumn{1}{c|}{30.88(0.98)}           & \multicolumn{1}{c|}{0.898($\uparrow$)}          & \multicolumn{1}{c|}{7.01(0.41)}           & \multicolumn{1}{c|}{{\textbf{41.56(1.10)}}}     & \multicolumn{1}{c|}{24.71(0.67)}           & 20.45(0.47)           \\ \midrule
\rowcolor{gray!10} \multicolumn{1}{l|}{Llama3.3 70B}          & \multicolumn{1}{c|}{\textbf{18.63(1.32)}}  & \multicolumn{1}{c|}{{\ul 51.26(1.52)}}     & \multicolumn{1}{c|}{{\ul 41.85(1.36)}}     & \multicolumn{1}{c|}{\textbf{32.27(1.10)}}  & \multicolumn{1}{c|}{\textbf{0.941($\uparrow$)}} & \multicolumn{1}{c|}{\textbf{8.98(0.32)}}  & \multicolumn{1}{c|}{{\ul 40.22(1.09)}}     & \multicolumn{1}{c|}{{\ul 31.43(0.64)}}     & {\ul 20.55(0.53)}     \\ \bottomrule
\multicolumn{10}{l}{Note: The best performance is \textbf{bolded} and the second performance is \underline{underlined}, the values in brackets are the variances.}                                                                                                                                                                                                                                                                                                                                                                                            
\end{tabular}
} 
\vspace{-10 pt}
\end{table*}

\subsection{Experimental Setting}

We selected a series of state-of-the-art open-source LLMs to evaluate our framework. The selection process was guided by publicly available benchmark rankings\footnote{Open LLM Leaderboard: \url{https://huggingface.co/spaces/open-llm-leaderboard/open_llm_leaderboard\#/?official=true}}. Since our goal is to evaluate the capability of CBR augmented different LLMs on the decision-making in the SCDSs,
eight LLMs with varying parameter sizes are finally chosen: Mistral 7B, Mistral-Nemo 12B, Mixtral 8x7B, Phi-4 14B, Deepseek-R1 14B, Gemma 2 27B, Qwen2.5 32B, and Llama 3.3 70B.
In our implementation, we leveraged the Ollama\footnote{Ollama:\url{https://ollama.com/}} to locally access the selected open-source models. Utilizing its API, we programmatically interacted with these running models, facilitating seamless integration of LLM capabilities into our applications.

To evaluate the decision-making results of evasive maneuvers, we used accuracy as a metric, comparing the LLM recommended maneuvers with the actual ones executed by human drivers. Moreover, we evaluated the quality of the generated text (Event context and ego-car maneuver justification) to assess interpretability and alignment with expert reasoning using BLEU4 \cite{papineni2002bleu}, METEOR \cite{banerjee2005meteor}, ROUGE-L \cite{lin2004automatic} and CIDEr \cite{vedantam2015cider} widely used in the NLP community. 
The ``event context'' output assesses the LLMs' comprehension of the SCDSs based on all input prompts, while the ``maneuver justification'' explains the rationale behind the contextually appropriate maneuver recommendation.

Experiments across various settings were conducted to answer the following three questions:
\begin{itemize}[leftmargin=15pt,itemsep=2pt,topsep=0pt,parsep=0pt]
\item \textbf{Q1}: \textit{What is the performance of different LLMs on the SCDS events?}
\item \textbf{Q2}: \textit{How do different prompt settings (risk-aware and risk-unaware) affect performance?}
\item \textbf{Q3}: \textit{Does augmenting LLMs with CBR lead to improved performance?}
\end{itemize}

\subsection{Vanilla LLM Maneuver Decision Performance Comparison (Q1)}
To assess the baseline capability of various LLMs in understanding and responding to SCDSs, we conducted a comprehensive comparison of eight models under a risk-unaware setting using all 1000 data (only feed the event caption into the task prompt with zero-shot).  As shown in the upper half of Table 3, overall \textit{Phi-4 14B, Deepseek-R1 14B, Gemma2 27B, Qwen2.5 32B, and Llama3.3 70B} consistently outperform other models across multiple dimensions. 

In the evasive maneuver recommendation task, which directly evaluates decision alignment with human drivers, \textit{Llama3.3 70B} significantly outperforms all other models with a micro-accuracy of 0.9369, followed closely by \textit{Deepseek-R1 14B} (0.9129) and \textit{Gemma2 27B} (0.8929). This trend suggests that larger models, particularly those with broader pretraining data and instruction-tuning, are better at recognizing subtle cues in risk driving scenes that inform safe decision-making.

While on the event context generation, which tests the models' capacity to understand the spatial-temporal context of risk events, \textit{Deepseek-R1 14B} achieves the highest BLEU4 (17.45) and CIDEr (30.55), while \textit{Qwen2.5 32B} and \textit{Llama3.3 70B} lead in METEOR (47.10) and ROUGE-L (46.24), respectively. These results indicate that while smaller models like \textit{Mistral-7B} and \textit{Mistral-Nemo} struggle with coherence and relevance, larger models exhibit superior capability in generating detailed and semantically accurate event narratives.
For the maneuver justification, which reflects the models' interpretability and ability to provide human-aligned rationales, \textit{Phi-4 14B} stands out in METEOR (38.20), while \textit{Deepseek-R1 14B} produces the highest ROUGE-L (38.52). Interestingly, \textit{Qwen2.5 32B} achieves the highest CIDEr score (30.11), despite relatively lower accuracy in maneuver prediction. This suggests that while some models may not always make the optimal decision, they can still offer rich, linguistically fluent justifications.

Overall, the latter five models consistently outperform smaller models across tasks (and hence in the following tests in Section 3.4 and 3.5, we evaluate our framework using these five high-performed LLMs). Their stronger language understanding, contextual reasoning, and decision alignment indicate their suitability for deployment in safety-critical driving assistance frameworks. Notably, Llama3.3 70B presents the most balanced and high-performing profile, excelling in both decision accuracy and narrative clarity, making it a strong candidate for integration into real-world systems. 

\subsection{Risk-Aware vs. Risk-Unaware Maneuver Decision Performance  (Q2)}

To evaluate the contribution of explicit risk modeling, we compare our framework's performance under two prompt configurations: risk-unaware and risk-aware settings. The risk-aware setting introduces structured risk annotations into the reasoning process, including scenario risk types, and event caption. This additional information aims to guide LLMs toward more informed, context-sensitive decisions. 
As shown in the lower half of Table 3, across all the five high-performed models, the risk-aware setting generally improves performance in maneuver decision alignment (except for \textit{Phi-4 14B} showing a slight drop). 

\textit{Llama 3.3 70B} achieves the highest micro-accuracy in both settings, increasing from 0.937 in the risk-unaware setting to 0.941 in the risk-aware setting. \textit{Deepseek-R1 14B} also shows consistent improvement, rising from 0.913 (unaware) to 0.914 (aware).
\textit{Qwen2.5 32B}, while slightly lower than Deepseek and Llama in performance, still improves from 0.875 to 0.898. \textit{Gemma2 27B} improves from 0.893 to 0.918, reflecting a substantial benefit from the risk-aware configuration. \textit{Phi-4 14B} shows a small decline in performance, as its risk-aware configuration unexpectedly underperforms in scenarios involving conflicts with adjacent vehicles.
Additional results in Table 4 and 5 on the scenario-specific analysis further confirm the robustness of risk-aware models. 

Overall, these results confirm that incorporating structured risk knowledge enhances both the factual grounding and interpretability of LLM-based decision-making in risk contexts. The improvement is especially evident in maneuver decision-making, where risk-aware models are more capable of capturing cause-effect dynamics and delivering reasoning that is more aligned with human decision making.

\begin{table*}[ht]
\setlength\tabcolsep{8pt}
\centering
\label{risk_unaware_scenarios}
\caption{Performance comparisons on risk-unaware evasive maneuver recommendation in various risk scenarios.}
\vspace{-10 pt}
\resizebox{0.9\linewidth}{!}{ %
\begin{tabular}{@{}lcccccccc@{}}
\toprule
\multicolumn{1}{c}{\multirow{2}{*}{\textbf{Risk scenario}}} & \multicolumn{8}{c}{\textbf{Risk-Unaware Model Micro-Accuracy}}                                                                               \\ \cmidrule(l){2-9} 
\multicolumn{1}{c}{}                               & Mistral 7B & Mistral-Nemo 12B & Mixtral 8x7B & Phi4 14B & Deepseek-R1 14B & Gemma2 27B   & Qwen2.5 32B     & Llama3.3 70B    \\ \midrule
Conflict with Vehicle Ahead                     & 0.8671     & 0.8741           & 0.8671       & 0.9441          & \textbf{0.9720} & 0.9161       & 0.9021          & {\ul 0.9650}    \\
Conflict with Oncoming Vehicle                    & 0.8112     & 0.8112           & 0.7762       & 0.7832          & 0.8182          & {\ul 0.8252} & 0.7762          & \textbf{0.8601} \\
Conflict with Adjacent Vehicle                    & 0.7343     & 0.8042           & 0.8042       & 0.8182          & {\ul 0.8531}    & 0.8252       & 0.7483          & \textbf{0.9161} \\
Conflict with Merging Vehicle                    & 0.8951     & 0.8881           & 0.8881       & \textbf{0.9510} & \textbf{0.9510} & {\ul 0.9301} & 0.9091          & \textbf{0.9510} \\
Head-on Conflict                                     & 0.8462     & 0.8042           & 0.8811       & 0.9161          & {\ul 0.9371}    & 0.9231       & \textbf{0.9510} & 0.9231          \\
Conflict with Opposite Turning vehicle                                   & 0.8322     & 0.8392           & 0.8741       & 0.8811          & {\ul 0.9161}    & 0.8881       & 0.8741          & \textbf{0.9580} \\
Conflict with Pedestrian                        & 0.9291     & 0.9291           & 0.9787       & 0.9645          & 0.9433          & 0.9433       & 0.9645          & \textbf{0.9858} \\ \midrule
\multicolumn{1}{c}{Overall}                        & 0.845     & 0.850           & 0.867       & 0.894          & {\ul 0.913}    & 0.893       & 0.875          & \textbf{0.937} \\ \bottomrule
\end{tabular}
}
\vspace{-5 pt}
\end{table*}

\begin{table*}[ht]
\setlength\tabcolsep{8pt}
\centering
\label{risk_aware_scenarios}
\caption{Performance comparisons on risk-aware evasive maneuver recommendation in various risk scenarios.}
\vspace{-10 pt}
\resizebox{0.68\linewidth}{!}{ %
\begin{tabular}{@{}lccccc@{}}
\toprule
\multicolumn{1}{c}{\multirow{2}{*}{\textbf{Risk scenario}}} & \multicolumn{5}{c}{\textbf{Risk-aware Model Micro-Accuracy}}                                  \\ \cmidrule(l){2-6} 
\multicolumn{1}{c}{}                               & Phi4 14b     & Deepseek-R1 14B & Gemma2 27B      & Qwen2.5 32B & Llama3.3 70B    \\ \midrule
Conflict with Vehicle Ahead                     & 0.9371               & {\ul 0.9441}         & 0.9301                 & 0.9371             & \textbf{0.9860}             \\
Conflict with Oncoming Vehicle                    & 0.7972               & 0.8112               & {\ul 0.8462}           & 0.7972             & \textbf{0.8731}             \\
Conflict with Adjacent Vehicle                    & 0.7483               & 0.8531               & {\ul 0.9021}           & 0.8392             & \textbf{0.9282}             \\
Conflict with Merging Vehicle                    & 0.9301               & {\ul 0.9510}         & {\ul 0.9510}           & 0.9091             & \textbf{0.9671}             \\
Head-on Conflict                                     & 0.9021               & \textbf{0.9301}      & \textbf{0.9301}        & 0.9161             & {\ul 0.9199}                \\
Conflict with Opposite Turning Vehicle                                   & 0.9161               & {\ul 0.9371}         & 0.9091                 & 0.9161             & \textbf{0.9388}             \\
Conflict with Pedestrian                        & 0.9645               & {\ul 0.9716}         & 0.9574                 & {\ul 0.9716}       & \textbf{0.9758}             \\ \midrule
\multicolumn{1}{c}{Overall}                        & 0.885($\downarrow$)       & 0.914($\uparrow$)          & {\ul 0.918($\uparrow$)}    & 0.898($\uparrow$)          & \textbf{0.941($\uparrow$)} \\ \bottomrule
\end{tabular}
}
\begin{minipage}[c]{0.3\textwidth}%
\centering
    \includegraphics[width=1\textwidth]{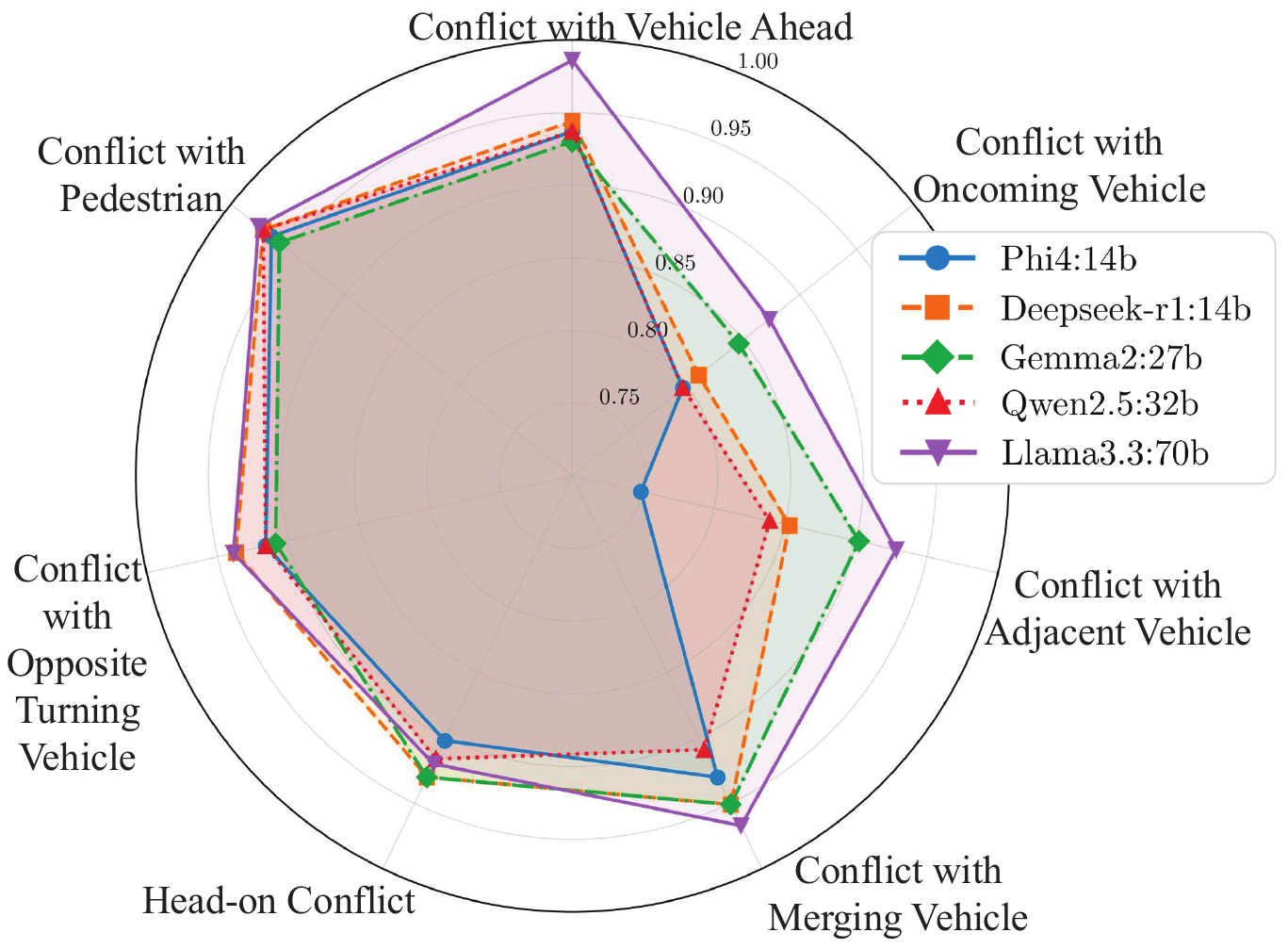}
\label{fig:figure}
\end{minipage}
\vspace{-25 pt}
\end{table*}

\subsection{CBR-Augmented Maneuver Decision Performance (Q3)}

\begin{figure*}[h]
\centering
\includegraphics[width=1.0\textwidth]{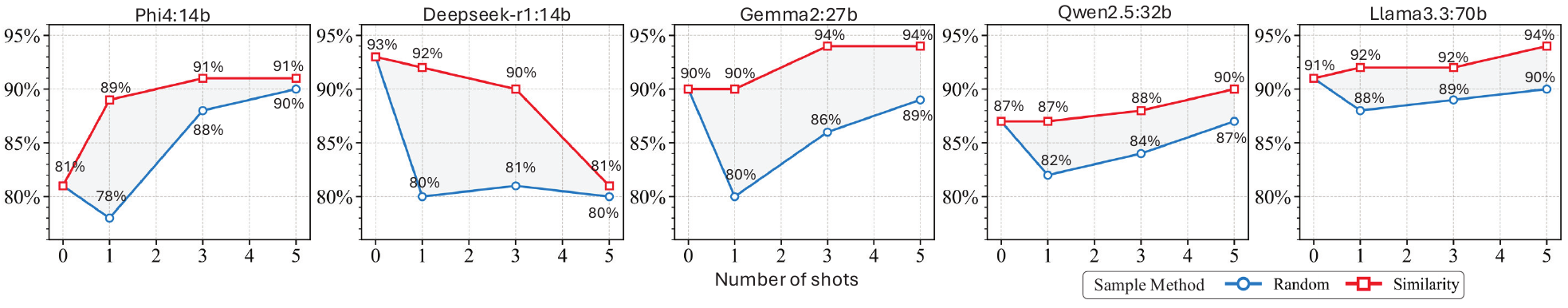}
\vspace{-10 pt}
\caption{Accuracy comparison of different models using two sampling methods across different shot settings.}
\label{few_shot_results_graph}
\vspace{-10 pt}
\end{figure*}

To test the performance of the CBR-augmented LLMs for driving maneuver recommendation, we conducted further experiments employing the five high-performed LLMS in Section 3.3 on new data usage plan. We randomly selected 100 samples balanced across seven risk scenarios (around 14 samples per types) that were used as test cases to assess the framework's generalization ability on unseen situations. The remaining 900 scenarios were stored in the case base, serving as the repository of past driving experiences leveraged by the CBR component. 
This data split allowed for a robust evaluation of the CBR-augmented LLMs' performance. 
The evaluation focuses on the impact of CBR-based few-shot prompting strategies, the number of in-context examples provided (1, 3, and 5 shots), and two distinct example selection methods: a random selection approach and a more informed similarity-based approach. The latter strategically selects past cases from the case base that belong to the same risk type as the current test scenario, leveraging similarity calculations to identify and retrieve the most relevant precedents for effective few-shot prompting. The results of this comprehensive evaluation are presented in Table 6 and Figure \ref{few_shot_results_graph}.

The accuracy trends shown in Figure \ref{few_shot_results_graph} demonstrate that the similarity-based method for selecting few-shot examples consistently yields superior performance across all evaluated LLMs compared to the random selection method. Notably, within the random selection strategy, we observed a general trend across all models where the introduction of a single, potentially irrelevant, example (1-shot) often led to a decrease in performance compared to the zero-shot baseline. This suggests that the reasoning capabilities of the LLMs can be misled by the inclusion of dissimilar cases in the prompt. However, as the number of randomly selected examples increased (3-shot and 5-shot), the performance generally improved, potentially due to the LLMs being able to discern a more general pattern from a larger, although still potentially noisy, set of examples. In contrast, the similarity-based method exhibited a more consistent and positive correlation between the number of few-shot examples and decision accuracy. An increasing number of relevant examples  (up to the tested limit of 5) in the prompt consistently enhanced the LLMs' ability to make accurate evasive maneuver decisions, underscoring the value of contextually aligned in-context learning.

However, the \textit{DeepSeek-R1} model presents a notable exception to this trend. As clearly visualized in Figure 4 and supported by the metrics in Table 6, increasing the number of few-shot examples consistently leads to a degradation in its performance, regardless of the selection method. 
This counter-intuitive behavior aligns with findings reported in the original \textit{DeepSeek-R1} paper \cite{guo2025deepseek}, indicating a unique characteristic where in-context learning via few-shot prompting negatively impacts its ability to effectively leverage the provided examples for this specific CBR-augmented driving maneuver decision-making task. This highlights the importance of model-specific evaluations when employing few-shot learning techniques.

\begin{table*}[ht]
\setlength\tabcolsep{10pt}
\centering
\label{few_shot_results}
\caption{Performance comparisons on CBR augmented evasive maneuver recommendation with different settings.}
\vspace{-10 pt}
\resizebox{\linewidth}{!}{ %
\begin{tabular}{c|c|ccccc|c|cccc}
\toprule
                                  &                                                                            & \multicolumn{1}{c|}{}                          & \multicolumn{4}{c|}{\textbf{Event Context}}                                                                                                                                                        & \textbf{Ego-car Evasive maneuver}              & \multicolumn{4}{c}{\textbf{Ego-car Maneuver Justification}}                                                                                                                                   \\ \cmidrule{4-12} 
\multirow{-2}{*}{\textbf{Model}}           & \multirow{-2}{*}{\begin{tabular}[c]{@{}c@{}}\textbf{Sample} \\ \textbf{Method}\end{tabular}} & \multicolumn{1}{c|}{\multirow{-2}{*}{\textbf{Shot}}}    & BLEU4                                        & METEOR                                       & ROUGE\_L                                     & CIDEr                                        & Micro-Average                         & BLEU4                                        & METEOR                                       & ROUGE\_L                                & CIDEr                                        \\ \midrule
                                  & -                                                                          & \multicolumn{1}{c|}{0}                         & 5.51(0.22)                                   & 34.56(0.82)                                  & 27.95(0.55)                                  & 19.73(0.31)                                  & 0.81                                  & 6.88(0.56)                                   & 37.69(1.52)                                  & 25.11(1.03)                             & 18.52(0.62)                                  \\ \cmidrule{2-12} 
                                  &                                                                            & \multicolumn{1}{c|}{1}                         & 3.16(0.09)                                   & 28.13(0.60)                                  & 24.58(0.40)                                  & 16.69(0.19)                                  & 0.78                                  & 5.02(0.25)                                   & 35.97(1.27)                                  & 24.84(0.71)                             & 16.92(0.32)                                  \\
                                  &                                                                            & \multicolumn{1}{c|}{3}                         & 8.26(0.61)                                   & 37.74(2.03)                                  & 29.76(1.08)                                  & 22.16(0.75)                                  & 0.88                                  & 7.06(0.63)                                   & 39.27(1.32)                                  & 25.90(1.50)                             & 19.04(0.66)                                  \\
                                  & \multirow{-3}{*}{Random}                                                   & \multicolumn{1}{c|}{5}                         & 11.64(0.52)                                  & 47.53(1.18)                                  & 32.55(0.83)                                  & 27.49(0.54)                                  & 0.90                                  & 4.64(0.18)                                   & 40.04(0.78)                                  & 20.76(0.63)                             & 17.93(0.33)                                  \\ \cmidrule{2-12} 
                                  & \cellcolor{gray!10}                                                   & \multicolumn{1}{c|}{\cellcolor{gray!10}1} & \cellcolor{gray!10}9.74(0.87)           & \cellcolor{gray!10}40.80(1.73)          & \cellcolor{gray!10}32.22(1.16)          & \cellcolor{gray!10}24.35(0.79)          & \cellcolor{gray!10}0.89          & \cellcolor{gray!10}7.33(0.54)           & \cellcolor{gray!10}40.78(1.19)          & \cellcolor{gray!10}27.29(1.26)     & \cellcolor{gray!10}19.27(0.54)          \\
                                  & \cellcolor{gray!10}                                                   & \multicolumn{1}{c|}{\cellcolor{gray!10}3} & \cellcolor{gray!10}14.12(1.01)          & \cellcolor{gray!10}{\ul 51.03(1.51)}      & \cellcolor{gray!10}37.24(1.22)          & \cellcolor{gray!10}28.99(0.88)          & \cellcolor{gray!10}0.91          & \cellcolor{gray!10}6.20(0.30)           & \cellcolor{gray!10}40.92(1.52)          & \cellcolor{gray!10}24.67(0.93)     & \cellcolor{gray!10}18.96(0.45)          \\
\multirow{-7}{*}{Phi4 14B}        & \multirow{-3}{*}{\cellcolor{gray!10}Similarity}                       & \multicolumn{1}{c|}{\cellcolor{gray!10}5} & \cellcolor{gray!10}14.06(0.75)          & \cellcolor{gray!10}49.82(1.26)          & \cellcolor{gray!10}36.67(0.95)          & \cellcolor{gray!10}28.90(0.69)          & \cellcolor{gray!10}0.91          & \cellcolor{gray!10}6.73(0.46)           & \cellcolor{gray!10}42.37(1.39)          & \cellcolor{gray!10}24.58(0.99)     & \cellcolor{gray!10}19.66(0.49)          \\ \midrule
                                  & -                                                                          & \multicolumn{1}{c|}{0}                         & {\ul 19.39(1.39)}                              & {\ul 51.17(1.90)}                              & {\ul 44.32(1.36)}                              & 31.97(1.22)                                  & {\ul 0.93}                            & 9.85(0.89)                                   & 39.79(2.11)                                  & 34.20(1.60)                             & 21.24(0.85)                                  \\ \cmidrule{2-12} 
                                  &                                                                            & \multicolumn{1}{c|}{1}                         & 2.61(0.07)                                   & 22.38(0.63)                                  & 22.79(0.43)                                  & 14.40(0.26)                                  & 0.80                                  & 5.50(0.37)                                   & 31.40(1.56)                                  & 26.34(0.88)                             & 15.36(0.40)                                  \\
                                  &                                                                            & \multicolumn{1}{c|}{3}                         & 10.70(1.41)                                  & 39.28(2.47)                                  & 35.89(1.62)                                  & 24.00(1.25)                                  & 0.81                                  & 9.30(0.86)                                   & 37.56(1.81)                                  & 31.82(1.69)                             & 20.09(0.94)                                  \\
                                  & \multirow{-3}{*}{Random}                                                   & \multicolumn{1}{c|}{5}                         & 8.51(0.68)                                   & 35.27(2.29)                                  & 32.68(1.27)                                  & 21.72(0.70)                                  & 0.80                                  & 7.41(0.66)                                   & 34.77(2.02)                                  & 31.15(1.55)                             & 18.54(0.78)                                  \\ \cmidrule{2-12} 
                                  & \cellcolor{gray!10}                                                   & \multicolumn{1}{c|}{\cellcolor{gray!10}1} & \cellcolor{gray!10}18.87(1.40)          & \cellcolor{gray!10}50.52(1.36)          & \cellcolor{gray!10}42.16(1.34)          & \cellcolor{gray!10}{\ul 32.07(1.15)}      & \cellcolor{gray!10}{\ul 0.92}    & \cellcolor{gray!10}8.21(0.76)           & \cellcolor{gray!10}37.29(1.99)          & \cellcolor{gray!10}31.21(1.72)     & \cellcolor{gray!10}19.54(0.79)          \\
                                  & \cellcolor{gray!10}                                                   & \multicolumn{1}{c|}{\cellcolor{gray!10}3} & \cellcolor{gray!10}18.78(1.61)          & \cellcolor{gray!10}48.71(2.00)          & \cellcolor{gray!10}{\ul 44.39(1.67)}      & \cellcolor{gray!10}31.39(1.26)          & \cellcolor{gray!10}0.90          & \cellcolor{gray!10}9.59(0.83)           & \cellcolor{gray!10}39.23(2.12)          & \cellcolor{gray!10}34.76(1.63)     & \cellcolor{gray!10}21.13(0.81)          \\
\multirow{-7}{*}{Deepseek-R1 14B} & \multirow{-3}{*}{\cellcolor{gray!10}Similarity}                       & \multicolumn{1}{c|}{\cellcolor{gray!10}5} & \cellcolor{gray!10}6.10(0.38)           & \cellcolor{gray!10}30.82(1.14)          & \cellcolor{gray!10}28.92(0.79)          & \cellcolor{gray!10}19.10(0.47)          & \cellcolor{gray!10}0.81          & \cellcolor{gray!10}8.49(0.73)           & \cellcolor{gray!10}36.78(1.80)          & \cellcolor{gray!10}31.23(1.54)     & \cellcolor{gray!10}19.45(0.85)          \\ \midrule
                                  & -                                                                          & \multicolumn{1}{c|}{0}                         & 10.03(1.04)                                  & 34.00(1.84)                                  & 38.51(1.14)                                  & 22.91(0.86)                                  & 0.90                                  & {\ul 10.20(0.58)}                              & 37.29(2.03)                                  & {\ul 36.37(1.80)}                         & {\ul 21.40(0.98)}                              \\ \cmidrule{2-12} 
                                  &                                                                            & \multicolumn{1}{c|}{1}                         & 3.80(0.23)                                   & 22.31(0.91)                                  & 29.59(0.82)                                  & 16.03(0.42)                                  & 0.80                                  & 6.41(0.66)                                   & 30.16(1.32)                                  & 33.82(2.12)                             & 17.56(0.87)                                  \\
                                  &                                                                            & \multicolumn{1}{c|}{3}                         & 5.87(0.31)                                   & 27.58(0.92)                                  & 33.71(0.84)                                  & 19.29(0.44)                                  & 0.86                                  & 7.43(0.72)                                   & 32.58(2.17)                                  & 32.91(2.37)                             & 18.29(0.92)                                  \\
                                  & \multirow{-3}{*}{Random}                                                   & \multicolumn{1}{c|}{5}                         & 9.99(1.26)                                   & 32.96(2.75)                                  & 36.26(1.77)                                  & 22.53(1.23)                                  & 0.89                                  & {\ul 10.10(0.97)}                              & 37.10(2.07)                                  & \textbf{37.94(2.31)}                    & {\ul 21.59(0.98)}                              \\ \cmidrule{2-12} 
                                  & \cellcolor{gray!10}                                                   & \multicolumn{1}{c|}{\cellcolor{gray!10}1} & \cellcolor{gray!10}9.38(0.98)           & \cellcolor{gray!10}34.68(2.28)          & \cellcolor{gray!10}37.44(1.33)          & \cellcolor{gray!10}23.40(1.09)          & \cellcolor{gray!10}0.90          & \cellcolor{gray!10}\textbf{11.20(1.29)} & \cellcolor{gray!10}37.73(2.14)          & \cellcolor{gray!10}{\ul 37.72(2.57)} & \cellcolor{gray!10}\textbf{22.20(1.25)} \\
                                  & \cellcolor{gray!10}                                                   & \multicolumn{1}{c|}{\cellcolor{gray!10}3} & \cellcolor{gray!10}17.74(1.38)          & \cellcolor{gray!10}46.45(1.75)          & \cellcolor{gray!10}44.90(1.59)          & \cellcolor{gray!10}31.91(1.36)          & \cellcolor{gray!10}\textbf{0.94} & \cellcolor{gray!10}7.17(0.53)           & \cellcolor{gray!10}36.59(1.26)          & \cellcolor{gray!10}28.04(1.13)     & \cellcolor{gray!10}18.93(0.58)          \\
\multirow{-7}{*}{Gemma2 27B}      & \multirow{-3}{*}{\cellcolor{gray!10}Similarity}                       & \multicolumn{1}{c|}{\cellcolor{gray!10}5} & \cellcolor{gray!10}{\ul 19.15(1.18)}      & \cellcolor{gray!10}47.03(1.59)          & \cellcolor{gray!10}{\ul 45.04(1.20)}      & \cellcolor{gray!10}{\ul 32.27(1.04)}      & \cellcolor{gray!10}\textbf{0.94} & \cellcolor{gray!10}8.28(0.67)           & \cellcolor{gray!10}37.70(1.46)          & \cellcolor{gray!10}30.34(1.32)     & \cellcolor{gray!10}19.73(0.68)          \\ \midrule
                                  & -                                                                          & \multicolumn{1}{c|}{0}                         & 15.98(0.75)                                  & 50.57(1.18)                                  & 39.78(1.05)                                  & 30.20(0.66)                                  & 0.87                                  & 5.62(0.24)                                   & 39.98(1.14)                                  & 22.81(0.67)                             & 18.76(0.43)                                  \\ \cmidrule{2-12} 
                                  &                                                                            & \multicolumn{1}{c|}{1}                         & 4.74(0.32)                                   & 28.92(0.86)                                  & 26.43(0.70)                                  & 17.29(0.44)                                  & 0.82                                  & 7.81(0.57)                                   & 36.01(1.44)                                  & 30.98(1.36)                             & 18.17(0.54)                                  \\
                                  &                                                                            & \multicolumn{1}{c|}{3}                         & 7.07(0.40)                                   & 34.45(1.14)                                  & 31.16(0.84)                                  & 20.30(0.45)                                  & 0.84                                  & 9.75(0.70)                                   & 40.73(1.74)                                  & 33.38(1.35)                             & 21.02(0.77)                                  \\
                                  & \multirow{-3}{*}{Random}                                                   & \multicolumn{1}{c|}{5}                         & 11.07(0.81)                                  & 39.56(1.84)                                  & 33.70(1.14)                                  & 24.08(0.76)                                  & 0.87                                  & 7.89(0.48)                                   & 38.21(1.38)                                  & 30.74(1.25)                             & 19.35(0.53)                                  \\ \cmidrule{2-12} 
                                  & \cellcolor{gray!10}                                                   & \multicolumn{1}{c|}{\cellcolor{gray!10}1} & \cellcolor{gray!10}11.39(1.16)          & \cellcolor{gray!10}40.71(1.56)          & \cellcolor{gray!10}35.26(1.17)          & \cellcolor{gray!10}24.69(0.95)          & \cellcolor{gray!10}0.87          & \cellcolor{gray!10}9.97(1.00)           & \cellcolor{gray!10}41.50(1.50)          & \cellcolor{gray!10}32.23(1.53)     & \cellcolor{gray!10}{\ul 21.43(0.89)}      \\
                                  & \cellcolor{gray!10}                                                   & \multicolumn{1}{c|}{\cellcolor{gray!10}3} & \cellcolor{gray!10}16.55(1.28)          & \cellcolor{gray!10}50.35(1.69)          & \cellcolor{gray!10}40.72(1.48)          & \cellcolor{gray!10}30.45(1.07)          & \cellcolor{gray!10}0.88          & \cellcolor{gray!10}7.52(0.34)           & \cellcolor{gray!10}42.51(1.42)          & \cellcolor{gray!10}28.12(1.16)     & \cellcolor{gray!10}20.12(0.42)          \\
\multirow{-7}{*}{Qwen2.5 32B}     & \multirow{-3}{*}{\cellcolor{gray!10}Similarity}                       & \multicolumn{1}{c|}{\cellcolor{gray!10}5} & \cellcolor{gray!10}18.14(1.57)          & \cellcolor{gray!10}50.54(1.62)          & \cellcolor{gray!10}41.91(1.52)          & \cellcolor{gray!10}31.16(1.15)          & \cellcolor{gray!10}0.90          & \cellcolor{gray!10}9.16(0.57)           & \cellcolor{gray!10}{\ul 43.42(1.77)}      & \cellcolor{gray!10}30.60(1.41)     & \cellcolor{gray!10}21.48(0.67)          \\ \midrule
                                  & -                                                                          & \multicolumn{1}{c|}{0}                         & 11.55(1.44)                                  & 39.58(2.51)                                  & 35.74(1.97)                                  & 24.98(1.27)                                  & 0.91                                  & 7.94(0.98)                                   & 41.55(1.68)                                  & 27.41(1.70)                             & 20.42(1.02)                                  \\ \cmidrule{2-12} 
                                  &                                                                            & \multicolumn{1}{c|}{1}                         & 5.99(0.37)                                   & 29.62(1.03)                                  & 29.68(0.90)                                  & 19.21(0.43)                                  & 0.88                                  & 5.85(0.37)                                   & 34.92(1.21)                                  & 25.91(0.93)                             & 17.68(0.49)                                  \\
                                  &                                                                            & \multicolumn{1}{c|}{3}                         & 7.50(0.60)                                   & 34.52(1.09)                                  & 33.00(1.08)                                  & 21.23(0.63)                                  & 0.89                                  & 7.51(0.55)                                   & 38.59(1.54)                                  & 29.16(1.24)                             & 19.21(0.61)                                  \\
                                  & \multirow{-3}{*}{Random}                                                   & \multicolumn{1}{c|}{5}                         & 14.06(1.53)                                  & 43.52(2.11)                                  & 38.64(1.82)                                  & 27.84(1.28)                                  & 0.90                                  & 8.07(0.59)                                   & {\ul 43.25(1.97)}                              & 29.52(1.01)                             & 20.76(0.68)                                  \\ \cmidrule{2-12} 
                                  & \cellcolor{gray!10}                                                   & \multicolumn{1}{c|}{\cellcolor{gray!10}1} & \cellcolor{gray!10}17.42(1.26)          & \cellcolor{gray!10}50.04(1.49)          & \cellcolor{gray!10}40.74(1.12)          & \cellcolor{gray!10}{\ul 32.08(1.07)}      & \cellcolor{gray!10}{\ul 0.92}    & \cellcolor{gray!10}4.74(0.18)           & \cellcolor{gray!10}39.33(1.13)          & \cellcolor{gray!10}18.07(0.51)     & \cellcolor{gray!10}18.80(0.47)          \\
                                  & \cellcolor{gray!10}                                                   & \multicolumn{1}{c|}{\cellcolor{gray!10}3} & \cellcolor{gray!10}18.58(1.23)          & \cellcolor{gray!10}49.15(1.62)          & \cellcolor{gray!10}44.41(1.02)          & \cellcolor{gray!10}31.85(0.99)          & \cellcolor{gray!10}{\ul 0.92}    & \cellcolor{gray!10}8.21(0.72)           & \cellcolor{gray!10}42.58(1.69)          & \cellcolor{gray!10}26.49(1.47)     & \cellcolor{gray!10}21.07(0.81)          \\
\multirow{-7}{*}{Llama3.3 70B}    & \multirow{-3}{*}{\cellcolor{gray!10}Similarity}                       & \cellcolor{gray!10}5                      & \cellcolor{gray!10}\textbf{21.81(1.63)} & \cellcolor{gray!10}\textbf{53.23(1.74)} & \cellcolor{gray!10}\textbf{47.17(1.42)} & \cellcolor{gray!10}\textbf{34.79(1.30)} & \cellcolor{gray!10}\textbf{0.94} & \cellcolor{gray!10}8.76(0.67)           & \cellcolor{gray!10}\textbf{43.44(1.44)} & \cellcolor{gray!10}26.83(1.29)     & \cellcolor{gray!10}{\ul 21.71(0.81)}      \\ \bottomrule
\end{tabular}
}
\vspace{-5 pt}
\end{table*}

\subsection{Case Study}

\begin{figure*}[htbp]
\centering
\begin{minipage}[t]{0.49\textwidth}
\centerline{\includegraphics[width=1\linewidth]{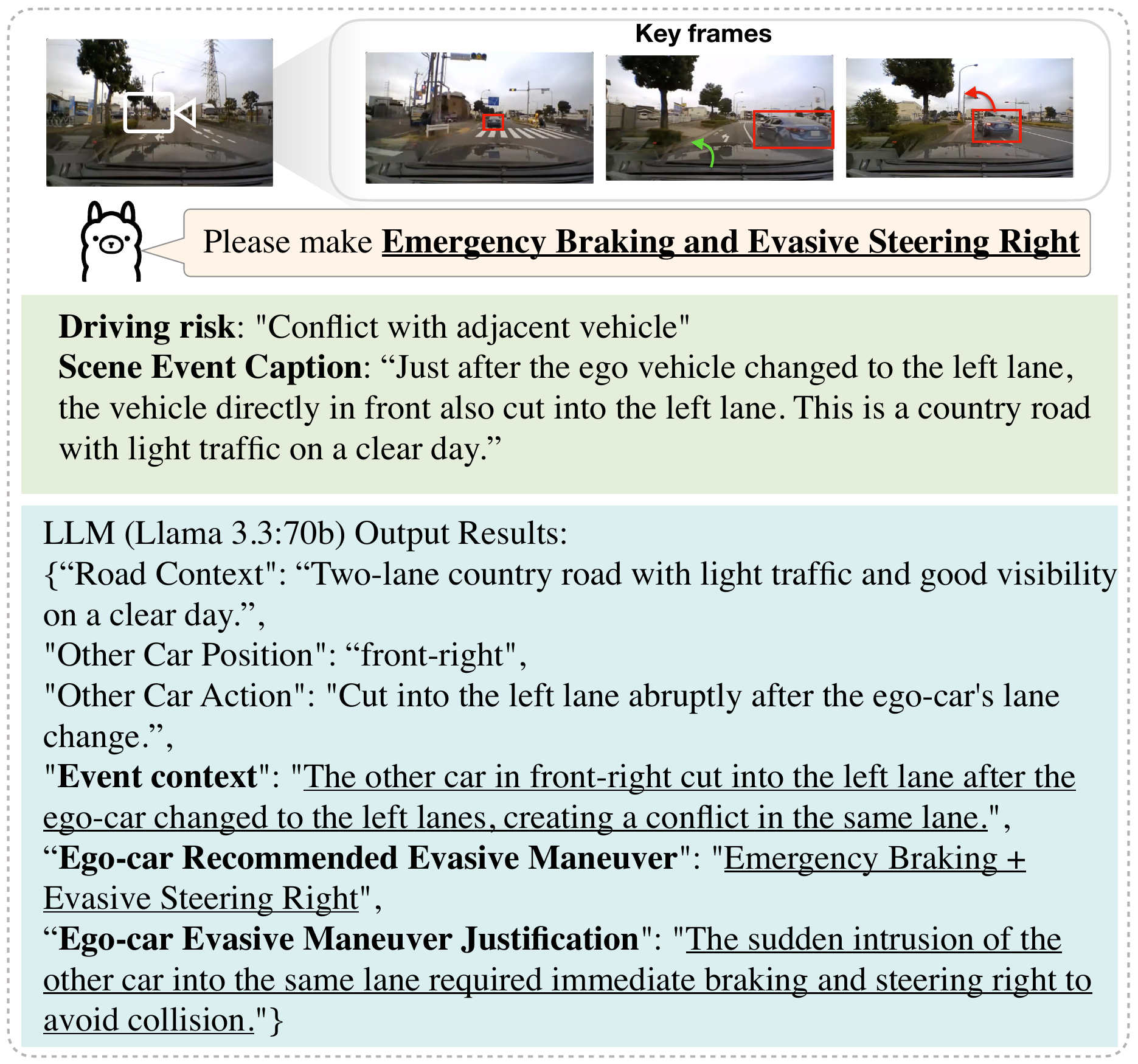}}
\vspace{-5 pt}
\caption{Case one for risk scenarios of ``conflict with a adjacent vehicle''.}
\label{example1}
\end{minipage}
\begin{minipage}[t]{0.49\textwidth}
\centerline{\includegraphics[width=1\linewidth]{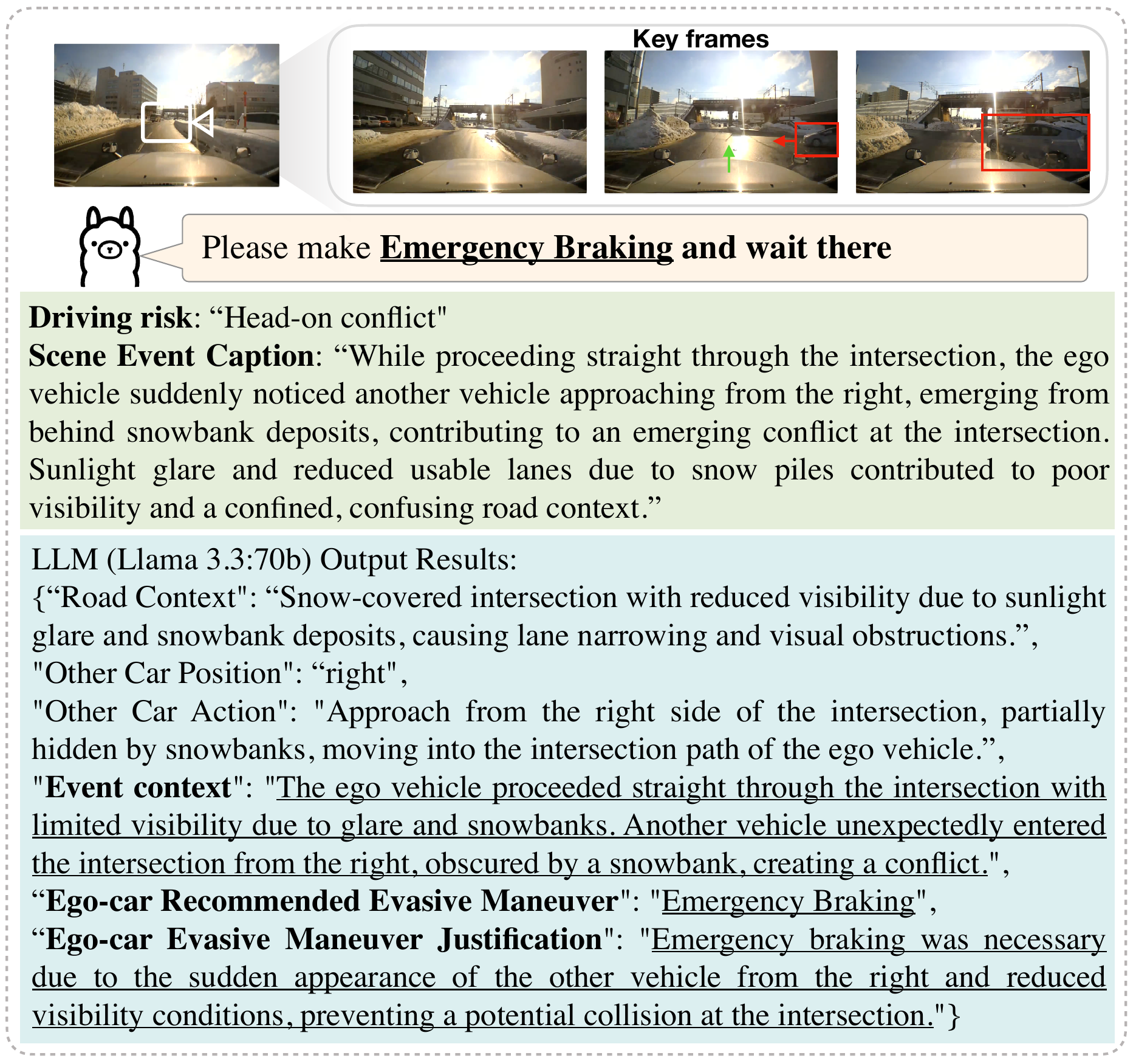}}
\vspace{-5 pt}
\caption{Case two for risk scenarios of ``head-on conflict''.}
\label{example2}
\vspace{-10 pt}
\end{minipage}
\end{figure*}

To further illustrate the practical application and effectiveness of our CBR-LLM framework, we present two case studies based on two different risk scenarios, as depicted in Figure \ref{example1} and \ref{example2}. 

In both cases, we employed the \textit{Llama 3 70B} model (identified as the best-performing model in Section 3.5) enhanced with our similarity-based CBR method for few-shot example retrieval. The model was tasked with recommending an appropriate evasive maneuver along with a justification. The LLM's output demonstrates a strong understanding of the critical situation.

The first case (Figure 5) involved a ``conflict with adjacent vehicle'' where the ego-vehicle, after a lane change, faced a sudden intrusion from a nearby vehicle. Our CBR-augmented Llama model accurately assessed the situation, recommending ``Emergency Braking and Evasive Steering Right'' with the clear justification.
The second case (Figure 6) depicted a ``Head-on conflict'' at a snow-covered intersection. Another vehicle unexpectedly approached from the right, partially obscured, creating an emerging conflict. The CBR-augmented Llama model correctly identified the situation. Its recommended ``Emergency Braking'' was a cautious and contextually appropriate response.

These two diverse case studies highlight the robustness and adaptability of our CBR-LLM framework. In both scenarios, the model demonstrated a strong ability to comprehend the risk driving event and generate situationally appropriate evasive maneuvers coupled with logical and coherent justifications. The framework's capacity to reason effectively in both lateral and longitudinal conflict scenarios underscores its potential as a valuable tool for enhancing the safety and reliability of intelligent driving systems.

\section{Conclusions}

This paper presented a CBR-augmented LLM framework for evasive maneuver decision-making in safety-critical driving scenarios. Through extensive experimentation, we demonstrated that augmenting LLMs with relevant prior experiences retrieved from a structured case base significantly improves their ability to recommend accurate and explainable evasive maneuvers. The use of dashcam video inputs brings the framework closer to real-world deployment by capturing risk scenarios from a driver-centric perspective, while similarity-based few-shot prompting ensures that model outputs are grounded in contextually relevant precedents. Our findings show consistent improvements across decision accuracy and justification quality, particularly under risk-aware settings. Notably, the framework proves effective across a range of risk types and environmental complexities. These results affirm the feasibility of combining LLMs with experience-based retrieval for driving decision support. Looking forward, this line of research opens promising directions for building adaptive and interpretable AI co-pilots in both autonomous and human-in-the-loop driving systems, with future efforts focusing on real-time risk case updates, continual learning, and deployment in closed-loop simulation and field trials.

\bibliographystyle{kr}
\bibliography{sample-base}
\balance

\appendix

\section{Prompt for Decision Making in SCDSs}
\begin{figure}[h]
\centering
\includegraphics[width=0.5\textwidth]{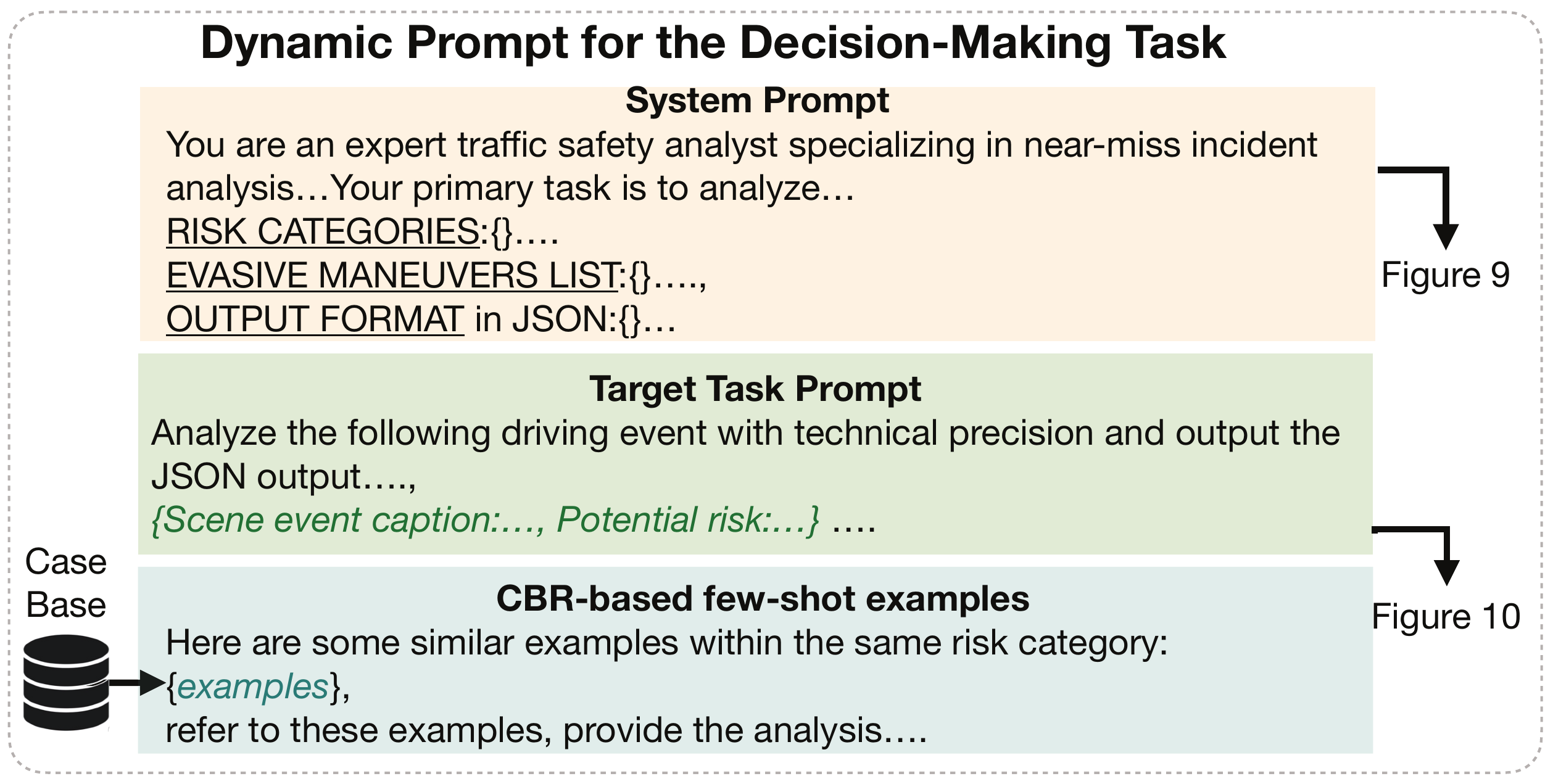}
\vspace{-10 pt}
\caption{Dynamic prompt setting for this task.}
\label{prompt_setting}
\vspace{-10 pt}
\end{figure}

Our prompt for this task is designed in three parts: system prompt, target task prompt and the CBR-based few-shot examples, as shown in Figure \ref{prompt_setting}. The system prompt is fixed for the entire task, it designates the LLMs working as an expert traffic safety analyst for near-miss incident analysis, introduces the primary task,  explains the definition of risk categories and list of ego-car response actions, and formats the final JSON output for LLM response (as shown in Figure 9). The target task prompt is designed specifically for the decision-making of a given driving event. Here the input is the target event related information (i.e., event id, event description and potential risk in the event), as shown in Figure 10. The LLMs are asked to make decision following a CoT process with additional analysis requirements. Here we join the CBR-based few-shot examples into the task prompt for few-shot experience. The dynamic examples provided to LLMs are retrieved from our evolving case base (with cases stored in specific format in the lower part of Figure 10). In our experiment, we compared different settings of case retrieval (random-based and similarity-based retrieval) and different numbers of cases as examples. 

\section*{Supplementary material}

Figure 9: System Prompts.

**SYSTEM PROMPTS**
```
"""
You are an expert traffic safety analyst specializing in near-miss incident analysis and risk
assessment. Your primary task is to accurately analyze driving event descriptions based on
the given risk categories, while extracting and inferring critical situational information.
You should conduct the technical analysis and provide the results in English.
Your analysis must be concise, accurate, and formatted in JSON strictly following the OUTPUT
FORMAT in JSON provided below. Use technical terminology consistently and maintain a clear
spatial reference framework.
{RISK CATEGORIES}
{Ego-Car Response}
{OUTPUT FORMAT in JSON}

Your response must include only the JSON output with no additional commentary or
explanations."""
```

**RISK CATEGORIES** (Refer to these definitions of risk categories when analyzing the events):
1. **Conflict with lead vehicle**: A situation where your vehicle (the ego vehicle) and the vehicle in front of you are at risk of colliding. This could happen if the lead vehicle stops suddenly, or if your vehicle fails to stop in time due to speed, following distance, or other factors like weather conditions.
2. **Conflict with oncoming vehicle**: A conflict arising when two vehicles traveling in opposite directions (on roads where traffic flows in both directions) are at risk of colliding, typically head-on. This could occur if one vehicle crosses over into the opposing lane for any reason, such as a loss of control or attempting to pass another vehicle.
3. **Conflict with adjacent vehicle**: A conflict between your vehicle and another vehicle traveling alongside you in an adjacent lane. It can happen due to merging, changing lanes without checking blind spots, or drifting out of one's lane.
4. **Conflict with merging vehicles**: Conflicts occurring when a vehicle is entering the flow of traffic from a side road, on-ramp, or parking lot and could potentially collide with your vehicle if either party fails to yield properly or misjudges the other's speed and distance.
5. **Conflict with turning vehicle in the intersection**: Conflicts involving situations where vehicles are turning at an intersection and could collide with your vehicle, especially if there's a misunderstanding of right-of-way rules, failure to yield when turning, or miscalculation of distances and speeds.
6. **Conflict with pedestrian**: Conflicts with pedestrians, particularly risky in intersections due to the higher volume of traffic and potential for turns. These conflicts often result from failure to yield to pedestrians by drivers or pedestrians failing to follow traffic signals.
7. **Head-on collision in the intersection**: A specific scenario where two vehicles approach an intersection from different directions (typically perpendicular to each other) and risk colliding head-on because they both proceed into the intersection at the same time, possibly due to misunderstandings of right-of-way rules or failure to stop at signals.

**Ego-Car Response** (choose exactly one based on the given risk category and situation):
- **"Emergency Braking"**: Decisively applying the brakes to rapidly reduce speed.
- **"Evasive Steering Left"**: Quickly steering the vehicle to the left to avoid an obstacle or hazard.
- **"Evasive Steering Right"**: Swiftly steering the vehicle to the right to evade a potential collision.
- **"Sudden Acceleration"**: Rapidly increasing speed to avoid a rear collision or other hazards behind.
- **"Emergency Braking + Evasive Steering Left"**: Brake firmly while steering left to avoid an obstacle ahead.
- **"Emergency Braking + Evasive Steering Right"**: Brake hard and steer right to prevent a collision.
- **"Acceleration + Evasive Steering Left"**: Increase speed while steering left, typically to bypass a hazard on the right side.
- **"Acceleration + Evasive Steering Right"**: Speed up and steer right to avoid an obstacle on the left

**OUTPUT FORMAT in JSON** (DO NOT MODIFY THIS STRUCTURE):
{
**"potential risk"**: [Use EXACTLY the risk category provided in the input],
**"Road Context"**: [Describe the specific road configuration, conditions, and environmental factors present in the event. Include details about lane configuration, weather conditions if mentioned, and any relevant road features],
**"Event Context"**: [Provide a detailed chronological sequence of how the risk situation develops, focusing on the interactions of involving vehicles and circumstances that lead to the event],
**"Other Car Position"**: [Specify the exact relative positioning of other vehicles to the ego vehicle, using clear directional terms (e.g., front, rear, left, right, front left)],
**"Other Car Action"**: [Describe the complete sequence of movements or maneuvers executed by the other vehicle, including any indicators or signals used],
**"Ego-Car Action"**: [Select exactly ONE of the eight standardized responses listed above that best matches the situation],
**"Ego Action Justification"**: [Provide a technical, clear and concise explanation of why the chosen ego-car action is the most appropriate response given the specific risk category and situation]
}

Figure 10: Task Prompts.

**RISK-AWARE TASK PROMPTS WITH CBR-based FEW-SHOT EXAMPLES:**

```
"""Create precise analysis results for this driving event with the given risk category
(Potential Risk) and output the specified format (OUTPUT FORMAT in JSON).
```
**Event ID:** {event_id}
**Event Description:** {event_text}
**Potential Risk:** {risk_category}

You should conduct your decision making following this process:
1. Given the Event Description and Potential Risk, first conducts an overview of the general road context;
2. Analyzes the risk-evolved other car's position and action in a spatial-temporal inference manner, and sum-marizes a clear event context;
3. finally infers the evasive maneuver based on all the obtained information, and gives its justification why this maneuver is recommended.
   <reasoning><reasoning><repeat until you have a decision>

Technical Analysis Requirements:
1. Use the provided risk category as the foundation for your analysis
2. Select ONE standardized emergency response that best addresses the given risk category
3. Extract and analyze all vehicle movements and positions with precision
4. Document the exact sequence of decisions and reactions
5. Evaluate spatial relationships and timing in detail
6. Consider environmental factors and road conditions that influence the risk situation

Here are some similar examples within the same risk category:
{**examples**}
Refer to these examples, provide the analysis results for the current event in the specified JSON format without additional commentary."""

---

**EXAMPLE FORMAT IN THE CASE DATABASE**

EXAMPLE {i}:
{
"Event ID": "204",
"Event Description": "While moving straight ahead, a vehicle suddenly cut into the front of ego-car from the left and sharply decelerated by applying emergency braking. This vehicle attempted to overtake a bus on the right but then changed lanes back to the left at an intersection to make a left turn, returning to behind the bus.",
"Risk Category": "Conflict with lead vehicle",
**"Analysis Results":**
  {
    "Road Context": "Straight road with multiple lanes. Intersection ahead where vehicles may turn left.",
    "Event context": "A vehicle cuts into the front of the ego-car from the left and sharply decelerates by applying emergency braking. The other vehicle was attempting to overtake a bus on the right but then changed lanes back to the left at an intersection to make a left turn, returning to behind the bus.",
    "Other Car Position": "Directly in front of the ego-car in the same lane after cutting into its lane from the left.",
    "Other Car Action": "Cut into the ego-car's lane from the left, applied emergency braking, attempted to overtake a bus on the right, then changed lanes back to the left at an intersection to make a left turn.",
    "Ego-Car recommended Action": "Emergency Braking + Evasive Steering Right",
    "Ego-Car Action Justification": "The ego-car needed to rapidly decelerate and adjust its position to avoid rear-ending the lead vehicle that suddenly cut in and braked sharply. Emergency braking was necessary to reduce speed quickly, while evasive steering right helped maintain safe spacing and avoid collision."}
  }
},
EXAMPLE {i+1}:
{........."Risk Category": "Conflict with lead vehicle",}.........



\end{document}